\documentclass[10pt,journal]{IEEEtran}

\usepackage{times}
\usepackage{epsfig}
\usepackage{graphicx}
\usepackage{amsmath}
\usepackage{amssymb}

\usepackage{hyperref}

\usepackage{color}
\usepackage{epsfig}
\usepackage{epstopdf}
\usepackage{subfloat}
\usepackage{subcaption}
\usepackage{lineno}
\usepackage{float}
\usepackage{units}
\usepackage{bm}
\usepackage{enumerate}
\usepackage{multirow}
\usepackage{rotating}
\usepackage[hang,flushmargin]{footmisc}
\usepackage{mdwlist}
\usepackage{multirow}
\usepackage{lipsum}
\usepackage{textcomp}
\usepackage[table,xcdraw]{xcolor}
\usepackage{tablefootnote}
\usepackage{threeparttable}
\usepackage{algorithm}
\usepackage{algpseudocode}

\usepackage[table,xcdraw]{xcolor}

\hypersetup{
	colorlinks=true,       % false: boxed links; true: colored links
	linkcolor=blue,          % color of internal links (change box color with linkbordercolor)
	citecolor=red,        % color of links to bibliography
	urlcolor=magenta           % color of external links
}
\begin{document}

%%%%%%%%% TITLE
%\title{River Ice Concentration Analysis with Deep Learning}
\title{River Ice Segmentation with Deep Learning}
\author{Abhineet Singh \hspace{1cm} Hayden Kalke \hspace{1cm} Mark Loewen \hspace{1cm} Nilanjan Ray \\
	University of Alberta\\
	{\tt\small asingh1,kalke1,mrloewen,nray1@ualberta.ca}
}

\maketitle
%\thispagestyle{empty}

%%%%%%%%% ABSTRACT
\begin{abstract}
	This paper deals with the problem of computing surface ice concentration for two different types of ice from digital images of river surface.
	It presents the results of attempting to solve this problem using several state of the art semantic segmentation methods based on deep convolutional neural networks (CNNs).
	This task presents two main challenges - very limited availability of labeled training data and presence of noisy labels due to the great difficulty of visually distinguishing between the two types of ice, even for human experts.
	The results are used to analyze the extent to which some of the best deep learning methods currently in existence can handle these challenges.
	The code and data used in the experiments are made publicly available to facilitate further work in this domain.

\end{abstract}

%%%%%%%%% BODY TEXT
\section{Introduction}
\label{sec_intro}

The study of surface ice concentration and variation over time and place is crucial for understanding the process of river ice formation.
The computation of temporal and spatial ice distributions can help to validate models of this process.
The additional ability to distinguish frazil ice from the sediment-carrying anchor ice can also increase the estimation accuracy of the sediment transportation capacity of the river.
Towards this end, a large amount of video data has been captured using UAVs and bridge mounted game cameras from two Alberta rivers during the winters of 2016 and 2017.
The objective of this work is to analyze this data and perform dense pixel wise segmentation on these images to automatically compute the concentrations of the two types of ice.

\begin{figure}[t]
	\includegraphics[width=0.48\textwidth]{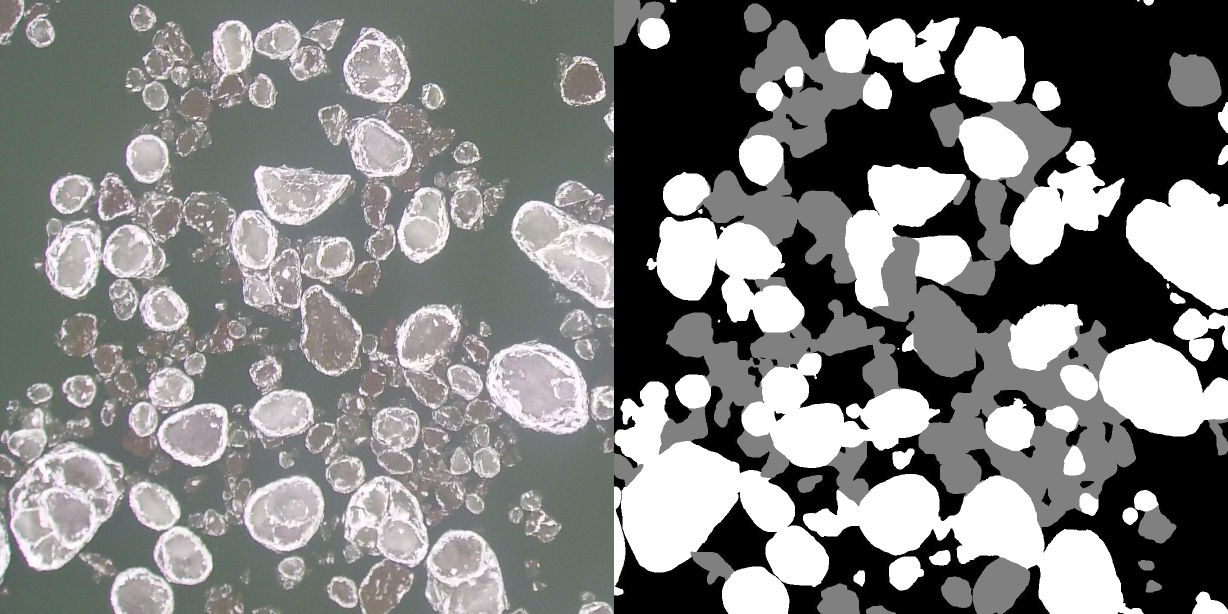}
	\caption{A sample training image with corresponding label where white, gray and black pixels respectively denote frazil ice, anchor ice, and water.}
	\label{fig_training_sample}
\end{figure}

The main challenge in this task is the lack of labeled data since it is extremely difficult and time consuming to manually segment images into the three categories, owing to the arbitrary shapes that the ice pans can assume.
As a result, there are currently only 50 labeled images (Fig. \ref{fig_training_sample}) to accompany 564 unlabeled test images and over 100 minutes of unlabeled high resolution (4K) videos.
These labeled images along with 205 additional images with only ice-water labeling had previously been used to train an SVM \cite{Kalke17_cripe,Kalke17_crst,Kalke17_thesis} to perform segmentation that provided water-ice classification accuracies ranging from $ 80.1\% $ -  $ 93.5 \% $ and surface ice concentration errors of $ 0.7\%$ - $3.3 \% $.
Though these methods were fairly successful at separating ice from water, they had difficulty in distinguishing between frazil and anchor ice pans, especially in cases where they are not physically separated and are hard to differentiate, even for human eyes.
This paper is mainly concerned with handling these more difficult cases.

To address the limitations of SVM-based ice classification, this work uses recent semantic segmentation methods based on deep CNNs.
Since CNNs need large amounts of training data to work well, several data augmentation techniques (Sec. \ref{sec_augmentation}) have been used to generate enough training images.
Detailed ablation studies have also been performed (Sec. \ref{sec_image_ablation},  \ref{sec_pixel_ablation}) to evaluate the extent to which few and partially labeled training images can be used to obtain results that are conducive to generating further training data with minimum human effort and thus setup a bootstrapping process.

\section{Background}
\label{sec_background}

Surface ice is formed on rivers in cold regions like Canada when the air temperature remains below freezing for extended periods of time.
It begins with the formation of frazil ice crystals and flocs as water columns become supercooled \cite{Daly94_frazil}. 
Being naturally adhesive \cite{Kempema11_frazil}, these ice crystals sinter together to form larger flocs whose increasing buoyancy brings them to the surface as slush.
As the moving slush is in turn subjected to freezing temperatures, it forms solid frazil pans and rafts which can become circular with upturned edges on colliding with each other.
Examples of these formations can be seen in Fig. \ref{fig_training_sample}.

Anchor ice is formed when frazil flocs come in contact with the river bed or other solid surface and freeze there to become immobile.
Such accumulations grow in multiple stages \cite{KERR2002101} to reach vertically towards the surface.
This mechanism increases the drag on the top of the accumulation which, along with internal buoyancy of the formation and the weakening effect of incoming solar radiation, can eventually overcome the bond between the formation and the substrate, causing it to break apart and float to the surface as anchor ice pans.
These pans carry entrapped sediments and other materials from the river bed and transport them downstream.

An improved understanding of the importance of such transportation to the overall sediment budget of the river would be useful for developing and validating models of river processes.
Thus, we require accurate estimates for the concentrations of sediment-carrying anchor ice in the river for which field measurements are unavailable.
Since there is far too much data for manual analysis to be practical, one of the objectives of this work is to estimate this concentration from digital images and videos of the river surface in an automated or semi-automated manner using deep learning.
%Another important application for river ice concentration estimation is to predict and prevent ice jamming and break up which can lead to flash floods, river bed modifications and bank scouring among other adverse consequences \cite{ANSARI20171}.
Distinguishing ice from water is relatively straightforward and has been accomplished fairly successfully using simple techniques like thresholding \cite{ANSARI20171} as well as classic machine learning methods like SVM with handcrafted features \cite{Kalke17_cripe,Kalke17_crst,Kalke17_thesis}.
The main goal of this work is therefore to be able to distinguish between frazil and anchor ice pans with high accuracy using state of the art deep learning techniques.

River ice images bear a significant resemblance to microscopic images of cells in the bloodstream which initially suggested the use of existing cell classification networks from medical imaging.
% after fine tuning them on the training images.
There are several promising studies employing a range of architectures including ConvNet \cite{Zhang17_convnet}, LeNet \cite{Shpilman17_lenet, Parthasarathy18_lenet}, Resnet \cite{Lei17_resnet} and Inception \cite{Li17_resnet_inception} that might have provided the base networks for such an approach.
However, a more detailed examination of these studies revealed that medical imaging tasks are mainly concerned with the detection and localization of specific kinds of cells rather than performing pixel wise segmentation that is necessary to estimate ice concentration.

Further, unsupervised and semi-supervised video segmentation techniques were considered to better utilize the large amount of unlabeled but high-quality video data that has been collected.
Most of these methods use optical flow for performing motion segmentation \cite{Faktor2014VideoSB}, though some appearance based \cite{Caelles17_osvos} and hybrid \cite{Grundmann10_graph, Jain17_fusionseg} methods have also been proposed.
An unsupervised bootstrapping approach has been proposed in \cite{pathakCVPR17learning} where motion segmented images are used as training data to learn an implicit representation of this object under the assumption that all moving pixels belong to the same foreground object.
This learnt model is then used to refine the motion segmentation and the improved results are in turn used to bootstrap further refinements.

Unfortunately, two underlying assumptions in \cite{pathakCVPR17learning}, and motion segmentation, in general, render such methods unsuitable for the current task.
First, they assume that there is a single moving foreground object whereas the objective here is to distinguish between two different types of moving ice, both of which are foreground objects.
Second, they assume a static background while the river, which makes up the background here, is itself moving.
Preliminary attempts to perform optical flow-based motion segmentation on the river ice videos confirmed its unsuitability for this work.
There is a method \cite{Tsai2016VideoSV} for performing simultaneous optical flow estimation and segmentation which might be able to address these limitations to some extent.
However, it had only a Matlab implementation available that was far too slow for our purpose so its further exploration was deferred to future work.

Finally, it seems that very limited existing work has been done on the application of deep learning for surface ice analysis as the only one that was found \cite{Junhwa17} uses microwave sensor data instead of images.
%and so is also not applicable to our scenario.
%I also found an existing work that uses deep learning for surface ice concentration estimation \cite{Junhwa17} but this utilizes microwave sensor data rather than images which made it inapplicable to our task.

\section{Methodology}
\label{sec_methodolgy}

\subsection{Data Collection and Labeling}
\label{sec_data_collection}

Digital images and videos of surface ice conditions were collected from two Alberta rivers - North Saskatchewan River and Peace River - in the 2016-2017 winter seasons.
Images from North Saskatchewan River were collected using both Reconyx PC800 Hyperfire Professional game cameras mounted on two bridges in Edmonton as well as a Blade Chroma UAV equipped 
with a CGO3 4K camera at the Genesee boat launch.
Data for the Peace River was collected using only the UAV at the Dunvegan Bridge boat launch and Shaftesbury Ferry crossing.
The game camera captured 3.1 megapixels resolution still images at one-minute frequency while the UAV camera captured 4K videos (8.3 megapixels) of up to 10 minutes duration.

Large $ 3840\times2160 $ UAV images were cropped into several $ 1280\times1080 $ images to make labeling more convenient while the smaller $ 2048\times1536 $ game camera images were only cropped to remove text information added by the camera software.
More than $ 200 $ of these images were labeled for binary ice-water classification but only $ 50 $ of these were labeled into 3 classes to distinguish between the two types of ice and the water.
Only the latter images were used for training in this work.
More details of the data collection and labeling process along with images of the camera setups are available in \cite[sec. 4.1]{Kalke17_crst}.

%This dataset provides digital images and videos of surface ice conditions were collected from two Alberta rivers - North Saskatchewan River and Peace River - in the 2016-2017 winter seasons.
%Images from North Saskatchewan River were collected using both Reconyx PC800 Hyperfire Professional game cameras mounted on two bridges in Edmonton as well as a Blade Chroma UAV equipped with a CGO3 4K camera at the Genesee boat launch.
%Data for the Peace River was collected using only the UAV at the Dunvegan Bridge boat launch and Shaftesbury Ferry crossing.
%The game camera captured 3.1 megapixels resolution still images at one-minute frequency while the UAV camera captured 4K videos (8.3 megapixels) of up to 10 minutes duration.
%
%It includes 50 manually labeled images with pixel-wise labels for 3 claases- anchor ice, frazil ice and water.
%Large 3840x2160 UAV images were cropped into several 1280x1080 images to make labeling more convenient while the smaller 2048x1536 game camera images were only cropped to remove text information added by the camera software.

\subsection{Image Segmentation}
\label{sec_segmentation}

% Table generated by Excel2LaTeX from sheet 'ice'
\begin{table}[t]
	\caption{Trainable parameter counts for the four models}
	\begin{tabular}{|l|l|l|l|l|}
		\hline
		\textbf{Model}      & \textbf{Deeplab} & \textbf{UNet} & \textbf{SegNet} & \textbf{DenseNet} \\ \hline
		\textbf{Parameters} & \scriptsize{41,050,787}     & \scriptsize{12,284,019}            & \scriptsize{11,546,739}     & \scriptsize{90,948}     \\ \hline
	\end{tabular}
	\label{tab_trainable_params}
\end{table}

Since neither cell classification nor video segmentation methods seemed promising, it was decided to rely only on supervised image segmentation.
After extensive research through several excellent resources for these methods \cite{awesom_segmentation,semantic_seg_dl}, four of the most widely cited and best performing methods with publicly available implementations were selected.
Descriptions of these methods that follow have been kept brief and high-level
because of the empirical nature of this work and the target audience.
%as being somewhat irrelevant to the empirical nature of this work and also of limited interest to its target readers.
%A fourth method based on a state of the art architecture, though not yet applied for segmentation, was also included.
%that I was able to get working in time.

The first of these models is UNet \cite{Ronneberger15_unet} from the medical imaging community.
It was introduced for neuronal structure segmentation in electron microscopic images and won the ISBI challenge 2015.
As the name suggests, UNet combines a contracting part with a symmetric expanding part to yield a U-shaped architecture that can both utilize contextual information and achieve good localization owing to the two parts respectively.
It was shown to be trainable with relatively few training samples while relying heavily on patch based augmentation which seemed to make it an ideal fit for this study.

The second network is called SegNet \cite{BadrinarayananK17_segnet} and was introduced for segmenting natural images of outdoor and indoor scenes for scene understanding application.
It uses a 13-layer VGG net \cite{SimonyanZ14a_vggnet} as its backbone and features a somewhat similar architecture as UNet.
The contracting and expanding parts are here termed as encoder and decoder, respectively and the upsampling units in the latter are not trainable, instead sharing weights with the corresponding max-pooling layers in the former.
Keras \cite{chollet2015keras} implementations were used for both UNet and SegNet, available as part of the same repository \cite{imgseg_keras}.
\footnote{This repository also includes two variants of the FCN architecture \cite{LongSD15_fcn,ShelhamerLD17_fcn} that were also tested but did not perform as well as the other two and are thus excluded here.}
%{\color{red} Is this important to say?}\\
%{\color{green} FCN methods are generally highly regarded in semantic segmentation literature and our present in many papers to represent the baseline.
%Since their implementation was readily available in the SegNet/Unet framework and I did test them quite thoroughly, including this line here might help to avoid reviewer questions about why we did not consider these.}

The third method is called Deeplab \cite{ChenPKMY18_deeplab} and is one of the best performing methods in the Tensorflow research models repository \cite{deeplab_repo}.
It uses convolutions with upsampled filters - the so called atrous convolutions \cite{Fisher16_atrous} - both to achieve better control over the feature response resolution and to incorporate larger context without increasing the computational cost.
%enlarge the field of view 
It also uses pyramidal max pooling to achieve scale-invariance and combines its last layer output with a fully connected conditional random field layer to improve localization accuracy while maintaining spatial invariance. 
This work uses a more recent version called Deeplabv3+ \cite{deeplabv3plus2018} which adds a decoder module to produce sharper object boundaries and uses the powerful Xception backbone architecture \cite{Chollet17_xception} for further performance improvements.

The fourth method is based on the DenseNet architecture \cite{HuangLMW17_densenet}.
To the best of our knowledge, this architecture has not yet been applied for segmentation but is included here due to its the desirable property of providing state of the art performance with a much smaller network size.
The basic idea of DenseNet is to connect each hidden layer of the network to all subsequent layers so that the feature maps output by each layer are used as input in all subsequent layers.
%and the total number of connections becomes quadratic in the number of layers.
This provides for better multi-resolution feature propagation and reuse while drastically reducing the total number of parameters and mitigating the vanishing gradient problem.
The architecture used in this work had 9 such layers though experiments were done with the more layers up to 21 (Sec. \ref{sec_augmentation}).
%with $ K = 640 $ and 18 with $ K=800 $ (Sec. \ref{sec_augmentation}) but there was no evidence of significant performance improvement.
As shown in Table \ref{tab_trainable_params}, DenseNet has by far the fewest parameters of the four models, being over two orders of magnitude smaller than the next smallest one.

%Indicator learning uses a loss function which forces the $ n $ channel output probability map (where $ n $ is the no. of classes) to learn one channel for each class by forcing the pairwise inner product between the probability values for labelled pixels to be 1 when the pixels belong to the same class and 0 otherwise.
%This has the disadvantage that, though it separates the pixels into distinct classes, it does not keep track of which channel corresponds to which class and hence class identification is not possible.
%In addition, it leads to a lot of flip-flopping during training as different channels are successively trained for different classes.
%To resolve these issues, the standard L2 loss was used instead.

\subsection{Data Augmentation and Training}
\label{sec_augmentation}

A simple sliding window approach was used to extract a large set of sub-images or patches from each training image.
The window was moved by a random stride between $ 10\% $ to $ 40\% $ of the patch size $ K $.
This process was repeated after applying random rotations to the entire image between 15 to 345 degrees divided into four bands of equal width to allow for multiple rotations for each image.
Finally, each patch was also subjected to horizontal and vertical flipping  to generate two additional patches.
All resultant patches were combined together to create the dataset for each $ K $.
For testing a model, patches of size $ K $ were extracted from the test image using a stride of $ K $, segmentation was performed on each patch and the results were stitched back to get the final result.

%\subsection{Training}
%\label{sec_training}

All models were trained and tested using patch sizes $ K\in \{256, 384, 512, 640, 800, 1000\} $.
DenseNet turned out to perform best with  $ K=800 $  while all other models did so with  $ K=640 $. 
All results in Sec. \ref{sec_results} were therefore obtained using these patch sizes.
The 50 labeled images were divided into two sets of 32 and 18 for generating the training and testing/validation images, respectively.
%Table \ref{tab_dataset_size} lists the sizes of the datasets thus generated for each patch size.
%Note that the training sets were used for generating the quantitative performance results on the validation sets while the combined sets generated using all $ 50 $ images
%(first column of Table \ref{tab_dataset_size})
%were used for producing qualitative results on the unlabeled validation set.
Results on some of the unlabeled videos (Sec. \ref{sec_qualitative_videos}) were also generated using models trained on all 50 images.

UNet and SegNet were both trained for 1000 epochs and the training and validation accuracies were evaluated after each.
The trained model used for testing was the one with either the maximum validation accuracy or the maximum mean accuracy depending on how well the training and validation accuracies were matched in the two cases.
%Table \ref{sec_augmentation} shows the test model accuracies for each $ K $.
Deeplab was trained for between $ 100, 000 $ and $ 200,000 $ steps.
Batch size of $ 10 $ was used for $ K=256 $ and $ 2 $ for $ K \in \{640, 800, 1000\} $ with the latter chosen due to memory limitations.
$ K=384 $ was tested with batch sizes $ 6 $ and $ 8 $ while $ K=512 $ was tested with $ 6 $ and $ 2 $.
Most tests were conducted using the default stride of 16 with corresponding atrous rates of $ [6, 12, 18] $ though one model with $ K=256 $ was also trained using  Stride $ 8 $ with atrous rates of $ [12, 24, 36] $.

DenseNet training was a bit more complicated.
Simply using all the pixels for training caused the network to rapidly converge to a model that labeled all pixels with the class that had the most training pixels - water in most cases.
To get meaningful results, the number of pixels belonging to each of the classes had to be balanced.
Therefore $ 10,000 $ random pixels belonging to each class were selected in each epoch, with different sets of pixels selected each time, and only these were used for computing the loss.
Training images with less than $ 10,000 $ pixels in any class were discarded.
%Also, DenseNet took much longer to train than the other networks and so was trained with a much smaller number of training images as shown in the last column of Table \ref{tab_dataset_size}.
The number of epochs were between $ 1000-1600 $ for all $ K $.
%except $ 1000 $ which could only be trained for $ 360 $ epochs due to time limitations.
In all cases, the performance metrics in Sec. \ref{sec_metrics} were computed on the validation set every 10 epochs and training was stopped when these became high enough
%(e.g. $ \texttt{pix\_acc} > 0.9 $)
or remained unchanged for over 100 epochs.

\subsection{Ablation Experiments}
\label{sec_testing}

One of the principal difficulties in training deep models for performing segmentation is the lack of sufficient labeled data due to the tedious and time-consuming manual segmentation of images.
This problem is exacerbated in the current task because of the difficulty in distinguishing between the two types of ice that exhibit both high intraclass variation and significant appearance overlap in addition to arbitrary and difficult to delineate shapes.
As a result, a highly desirable attribute of a practically applicable model would be its ability to learn from a few images including partially labeled ones.

Two different types of ablation experiments were performed in order to explore the suitability of the tested models in this regard.
The first one was to train the models using different subsets of the training set.
The second one was to consider the labels from only a small subset of pixels in each image to simulate the scenario of partially labeled training data.
Note that the input image itself was left unchanged so that the models did have access to all the pixels but the loss function minimized during training was computed using  only the labels from the selected pixels.

SegNet exhibited similar performance patterns as UNet in the ablation tests, while being slightly worse on average, probably because they share the same base network.
SegNet results have thus been excluded in Sec. \ref{sec_image_ablation} and \ref{sec_pixel_ablation} for the sake of brevity.
All the code and data used for training, augmentation and ablation tests is \textbf{made publicly available} \cite{Singh19_river_ice_segm_github}\cite{ebax-1h44-19} along with detailed instructions for using it to facilitate further work in this domain and easy replication of results.

\section{Results}
\label{sec_results}

%\begin{figure*}[!htbp]
%
%	\caption{
%		Results of ablation tests on  (left) UNet and  (right) DenseNet using 4, 8, 16, 24 and 32 training images.
%		Accuracy and IOU results are respectively shown with solid and dotted lines.
%	}
%	\label{fig_ablation_unet_densenet}
%\end{figure*}

%\begin{figure*}[!htbp]
%	\includegraphics[width=0.49\textwidth]{images/ablation_anchor}
%	\includegraphics[width=0.49\textwidth]{images/ablation_frazil}
%	\caption{
%		Comparing ablation performance using (left) anchor ice and (right) frazil ice.
%		Accuracy and IOU results are respectively shown with solid and dotted lines.
%		Note the different y-axis limits on the two plots.
%	}
%	\label{fig_ablation_img}
%\end{figure*}

\subsection{Evaluation Metrics}
\label{sec_metrics}

Following evaluation metrics are typically used in image segmentation \cite{ShelhamerLD17_fcn,img_seg_eval}:
\begin{enumerate}
	\item Pixel accuracy:
		\begin{align}
		\label{eq_pix_acc}
		\texttt{pix\_acc}
		=
		\frac{\sum_i n_{ii}}{\sum_i t_{i}}
		\end{align}
	\item Mean accuracy:
		\begin{align}
		\label{eq_mean_acc}
		\texttt{mean\_acc}
		=
		\frac{1}{n}
		\sum_{i} \frac{n_{ii}}{t_i}
		\end{align}
	\item Mean IOU:
		\begin{align}
		\label{eq_mean_iou}
		\texttt{mean\_iou}
		=
		\frac{1}{n}
		\sum_{i} \frac{n_{ii}}{t_i + \sum_{j}n_{ji} - n_{ii}}
		\end{align}
	\item Frequency Weighted IOU:
		\begin{align}
			\label{eq_fw_iou}
			\texttt{fw\_iou}
			=
			\sum_{k} (t_k)^{-1}	
		\sum_{i} \frac{t_i n_{ii}}{t_i + \sum_{j}n_{ji} - n_{ii}}
		\end{align}
\end{enumerate}
where $ n $ is the number of classes, $ n_{ij} $ is the number of pixels of class $ i $ predicted to belong to class $ j $ and  $ t_i $ is the total number of pixels of class $ i $ in the ground truth.
Note that accuracy measures only the rate of true positives  while IOU also accounts for false positives.
Hence, \textbf{accuracy and IOU are respectively equivalent to recall and precision} that are more widely used in pattern recognition and are referenced thus in the remainder of this paper.
Note also that {\small\texttt{pix\_acc}} is a frequency weighted version of {\small\texttt{mean\_acc}} 
%(just as {\small\texttt{fw\_iou}} is that of {\small\texttt{mean\_iou}})
and is thus referred to as \textbf{frequency weighted (fw) recall} in Table \ref{tab_acc_iou_32}.

Using these metrics to measure the combined segmentation performance over all three classes can lead to biased results when the number of image pixels is not evenly distributed between the classes.
This is particularly so in the current work whose main objective is to distinguish between the two types of ice.
% as it is almost trivial to separate water from ice.
However, as shown in Table \ref{tab_pixel_class_distribution}, more than half the pixels in both sets of test images are of water while anchor ice, which is the most difficult to segment, covers only about $ 17\% $ of the pixels.
Therefore, results in the next section are mostly restricted to class specific versions of these metrics.
%for each type of ice except in Fig. \ref{fig_acc_iou_32}.
%Combined results over ice and water are been restricted to .

% Table generated by Excel2LaTeX from sheet 'ice'
%\begin{table}[htbp]
%	\centering
%	\caption{Distribution of pixels in test images between the 3 classes}
%	\begin{tabular}{rrrr}
%		\multicolumn{1}{l}{\textbf{n\_test\_images}} & \multicolumn{1}{l}{\textbf{water}} & \multicolumn{1}{l}{\textbf{anchor ice}} & \multicolumn{1}{l}{\textbf{frazil ice}} \\
%		\textbf{18} & 0.5573 & 0.17479 & 0.26792 \\
%		\textbf{46} & 0.61019 & 0.16567 & 0.22414 \\
%	\end{tabular}%
%	\label{tab_pixel_class_distrivution}%
%\end{table}%

\begin{table}[!htbp]
	\centering
	\caption{Class frequencies in the two test sets}
	\begin{tabular}{|l|l|l|l|}
		\hline
		\textbf{test images} & \textbf{water} & \textbf{anchor ice} & \textbf{frazil ice} \\ \hline
		\textbf{18}              & 0.557    & 0.175          & 0.268          \\ \hline
		\textbf{46}              & 0.610     & 0.166          & 0.224          \\ \hline
	\end{tabular}
\label{tab_pixel_class_distribution}
\end{table}

In addition to these segmentation metrics, the estimated ice concentration accuracy has also been used.
This is computed through a three-step process.
First, column-wise ice concentration is obtained for each frame by computing the percentage of all pixels in each column that are classified as ice (combined, anchor or frazil).
These values are stacked together for all columns from left to right to form a vector of the same size as the frame width.
%, which is $ 3840 $ for the 4K videos used here.
Next, the mean absolute error (\textbf{MAE}) is computed between the ice concentrations vectors produced by the ground truth and a model to obtain the overall accuracy for that frame.
Finally, the median of MAE values over the entire test set is taken as the final metric.
Median has been preferred over mean as being significantly more robust to outliers.

The above metrics are only applicable to labeled images so that unlabeled videos can only be evaluated qualitatively but the conclusions thus obtained are usually somewhat subjective.
In order to ameliorate this, an unsupervised metric named \textbf{mean ice concentrations difference} has been proposed to measure the consistency of segmentation results between consecutive video frames as a proxy for its accuracy.
This metric is computed as the mean absolute difference between the ice concentration vectors of each pair of consecutive frames in the video.
Segmentation consistency over the entire video is summarized by taking the average of these  differences over all pairs of consecutive frames.
The intuition behind this metric is that, since the ice and/or river are moving slowly and video FPS is fairly high, the ice concentration changes very gradually and its difference between consecutive frames remains small.
A model that fails to generalize well to the videos would give inconsistent results in corresponding patches from nearby frames that would therefore result in a high mean concentration difference.
Experiments were also done using the direct pixel-wise difference between the segmentation masks themselves, both with and without incorporating motion estimation by optical flow \cite{Bouguet00_lk_opt_flow,Gunnar03_opt_flow,Ilg17_flownet2}, but the results were less consistent so these were excluded here.

\subsection{Quantitative Results}
\label{sec_quantitative}

\begin{table*}[t]
	\large
	\centering
	\caption{Segmentation recall and precision for SVM and all deep models trained and tested on the 32 and 18 image sets respectively.	
		The \texttt{fw} in the last category stands for frequency weighted and the corresponding recall and precision metrics refer to {\small\texttt{pix\_acc}} (Eq. \ref{eq_pix_acc}) and {\small\texttt{fw\_iou}} (Eq. \ref{eq_fw_iou}) as detailed in Sec. \ref{sec_metrics}.
		Relative increase over SVM is computed	as $ (\texttt{model\_value} - \texttt{svm\_value})/\texttt{svm\_value}\times 100 $
	}
	\begin{tabular}{|c|m{1.3cm}|m{1.54cm}|m{1.3cm}|m{1.54cm}|m{1.3cm}|m{1.54cm}|m{1.3cm}|m{1.54cm}|}
		\hline
		& \multicolumn{2}{c|}{\textbf{anchor ice}}                & \multicolumn{2}{c|}{\textbf{frazil ice}}                & \multicolumn{2}{c|}{\textbf{ice+water}}                 & \multicolumn{2}{c|}{\textbf{ice+water (fw)}}                                             \\ \cline{2-9} 
		\multirow{-2}{*}{\textbf{Model}} & {\normalsize\textbf{Metric Value (\%)}} &{\normalsize\textbf{Relative increase over SVM}}         & {\normalsize\textbf{Metric Value (\%)}} & {\normalsize\textbf{Relative increase over SVM} }        & {\normalsize\textbf{Metric Value (\%)}} & {\normalsize\textbf{Relative increase over SVM}}         & {\normalsize\textbf{Metric Value (\%)}} & {\normalsize\textbf{Relative increase over SVM}}          \\ \hline
				\multicolumn{9}{|c|}{\Large\textbf{Recall }}                                                                                                                                                                                                                                                                                                                                                                                                                                     \\ \hline
		\textbf{svm}                     & 61.54           & {-}           & 75.41           & -                                  & 78.12           & -                                  & 84.93           &              -                          \\ \hline
		\textbf{deeplab}                 & 74.46           & {\color[HTML]{009901} 21.00}          & \textbf{87.51}  & {\color[HTML]{009901} \textbf{16.05}} & \textbf{86.38}  & {\color[HTML]{009901} \textbf{10.57}} & \textbf{90.87}  & {\color[HTML]{009901} \textbf{7.00}}  \\ \hline
		\textbf{unet}                    & 73.75           & {\color[HTML]{009901} 19.85}          & 84.27           & {\color[HTML]{009901} 11.75}          & 85.13           & {\color[HTML]{009901} 8.97}           & 88.69           & {\color[HTML]{009901} 4.42}           \\ \hline
		\textbf{densenet}                & 76.96           & {\color[HTML]{009901} 25.06}          & 71.06           & {\color[HTML]{FE0000} -5.77}          & 81.42           & {\color[HTML]{009901} 4.22}           & 85.02           & {\color[HTML]{009901} 0.11}          \\ \hline
		\textbf{segnet}                  & \textbf{82.31}  & {\color[HTML]{009901} \textbf{33.75}} & 68.99           & {\color[HTML]{FE0000} -8.51}          & 83.06           & {\color[HTML]{009901} 6.32}           & 85.90           & {\color[HTML]{009901} 1.14}           \\ \hline
		\multicolumn{9}{|c|}{\Large\textbf{Precision }}                                                                                                                                                                                                                                        \\ \hline
%		& \multicolumn{2}{c|}{\textbf{anchor ice}}                   & \multicolumn{2}{c|}{\textbf{frazil ice}}                   & \multicolumn{2}{c|}{\textbf{ice+water}}                    & \multicolumn{2}{c|}{\textbf{ice+water (fw)}}              \\ \cline{2-9}
%		\multirow{-2}{*}{\textbf{Model}} & \textbf{Precision} & \textbf{\% increase over SVM}         & \textbf{Precision} & \textbf{\% increase over SVM}         & \textbf{Precision} & \textbf{\% increase over SVM}         & \textbf{Precision} & \textbf{\% increase over SVM}        \\ \hline
		\textbf{svm}                     & 43.32              & -                                  & 63.07              & -                                  & 65.84              &                 -                      & 76.84              & -                                 \\ \hline
		\textbf{deeplab}                 & \textbf{62.39}     & {\color[HTML]{009901} \textbf{44.03}} & \textbf{77.14}     & {\color[HTML]{009901} \textbf{22.32}} & \textbf{77.25}     & {\color[HTML]{009901} \textbf{17.33}} & \textbf{84.32}     & {\color[HTML]{009901} \textbf{9.72}} \\ \hline
		\textbf{unet}                    & 54.89              & {\color[HTML]{009901} 26.72}          & 71.17              & {\color[HTML]{009901} 12.84}          & 73.19              & {\color[HTML]{009901} 11.17}          & 81.73              & {\color[HTML]{009901} 6.36}          \\ \hline
		\textbf{densenet}                & 48.98              & {\color[HTML]{009901} 13.07}          & 60.97              & {\color[HTML]{FE0000} -3.32}          & 67.69              & {\color[HTML]{009901} 2.82}           & 77.49              & {\color[HTML]{009901} 0.84}          \\ \hline
		\textbf{segnet}                  & 52.80              & {\color[HTML]{009901} 21.90}          & 62.60              & {\color[HTML]{FE0000} -0.73}          & 69.60              & {\color[HTML]{009901} 5.72}           & 78.46              & {\color[HTML]{009901} 2.10}          \\ \hline	
	\end{tabular}
	\label{tab_acc_iou_32}
\end{table*}

%\begin{figure*}[!htbp]
%	%	{\centering\textbf{Your title}}	
%	%	\includegraphics[width=0.49\textwidth]{images/acc_32}
%	%	\includegraphics[width=0.49\textwidth]{images/iou_32}
%	\includegraphics[width=0.49\textwidth]{images/rec}
%	\includegraphics[width=0.49\textwidth]{images/prec}
%	\caption{
%		%		Performance of SVM  (dotted line) and all deep models (solid lines) on the 18-image test set using training with 32 images.
%		Performance of SVM and all deep models trained and tested on the 32 and 18 image sets respectively.	
%		The \texttt{fw} in the last legend entry stands for frequency weighted and the corresponding recall and precision metrics refer to {\small\texttt{pix\_acc}} (Eq. \ref{eq_pix_acc}) and {\small\texttt{fw\_iou}} (Eq. \ref{eq_fw_iou}) as detailed in Sec. \ref{sec_metrics}.
%		Note that Y-axis limits on the two plots are different to maximize line separability.
%	}
%	\label{fig_acc_iou_32}
%\end{figure*}

\subsubsection{Overview}
\label{sec_summary}

As shown in Table \ref{tab_acc_iou_32}, 
%shows the results for SVM and all the deep models trained and tested on the 32 and 18 image sets respectively.
all of the deep models provide significant improvement over SVM for all cases except a couple instances of frazil ice.
This is most notable for anchor ice where an increase of $ 12-20 \% $ in recall and $ 6-20\% $ in precision is achieved in absolute terms, with the respective relative increases being $ 19-34 \% $ and $ 13-44\% $.
It is noteworthy that the two best models – Deeplab and UNet – provide greater performance improvement, in both absolute and relative terms, with respect to precision than recall over all 4 categories.
This is particularly impressive since high precision is usually harder to achieve than recall, as testified by its lower values across all models and categories.

Further, we see that DenseNet and SegNet fall slightly behind SVM on frazil ice, especially with respect to recall, even though they have the two highest recalls on anchor ice.
This trend of an inverse relationship between anchor and frazil ice recall was consistently observed in the ablation tests too (Sec. \ref{sec_image_ablation}).
It seems that learning to better distinguish anchor from frazil ice often comes at the cost of either a decrease in the capability to recognize frazil ice itself or an overcorrection which causes some of the more ambiguous cases of frazil to be misclassified as anchor ice.
It is likely that the loss function can be minimized equally well by over-fitting either to frazil or to anchor ice, thus leading to two stable training states.

Comparing between the deep models themselves, Deeplab turns out to be the best overall followed closely by UNet.
%The corresponding dip in the frazil ice recall of these latter models, to a point that it is even slightly worse than SVM, is consistent with the inverse relationship between anchor and frazil ice performance mentioned above.
It is interesting to note that, while DenseNet and SegNet provide better recall with anchor ice, their corresponding precision is lower.
%A possibility is that both of these models, especially DenseNet, have a somewhat grainy quality in their outputs which is penalized more heavily by IOU than accuracy
Since recall does not penalize false positives while precision does, this probably indicates that DenseNet and SegNet misclassify frazil as anchor ice more often than Deeplab and UNet, which is consistent with the inverse relationship hypothesis.
Further, it can be seen that the performance difference between deep models and SVM decreases when all three classes are considered and even more so when the averaging is frequency weighted.
As mentioned before, these are the cases where high segmentation accuracy of water starts to dominate.
Finally, the greater difficulty of recognizing anchor ice over frazil ice is confirmed by its significantly lower recall and precision in almost all cases.

Table \ref{tab_mae} shows the ice concentration estimation accuracy of all the models in terms of median MAE.
It should be noted that recall and precision are better indicators of raw segmentation performance since MAE suffers from an averaging effect where false positives can cancel out false negatives to provide an overall concentration value that happens to be closer to the ground truth. 
As a result, MAE does not always provide a true indication of the actual recognition ability of the model.
As an example, SegNet has slightly better MAE than UNet on frazil ice  even though it has significantly lower ($ 9-16 \% $) precision and recall.
%exhibit the same performance patterns as  provided by the deep models as clearly as recall and precision.
Nevertheless, except for the single case of DenseNet with frazil ice, the deep models are consistently better than SVM here too, especially on anchor ice.
As with the segmentation metrics, Deeplab is the best model and provides around $ 3\% $ absolute and $ 40\% $ relative improvement over SVM.

\begin{table}[]
	\large
	\centering
	\caption{Median MAE for anchor and frazil ice over the 18 test images.
		Relative decrease over SVM is computed	as $ (\texttt{svm\_mae} - \texttt{model\_mae})/\texttt{svm\_mae}\times 100 $
	}
	\begin{tabular}{|m{1.4cm}|m{0.96cm}|m{1.54cm}|m{0.96cm}|m{1.54cm}|}
		\hline
		\multicolumn{1}{|c|}{}                                 & \multicolumn{2}{c|}{\textbf{anchor ice}}                                   & \multicolumn{2}{c|}{\textbf{frazil ice}}                                   \\ \cline{2-5} 
		\multicolumn{1}{|c|}{\multirow{-2}{*}{\textbf{Model}}} & {\normalsize\textbf{Median MAE (\%)}} & {\normalsize\textbf{Relative decrease over SVM}} & {\normalsize\textbf{Median MAE (\%)}} & {\normalsize\textbf{Relative decrease over SVM}} \\ \hline
		\textbf{svm}                                           & 8.37                & {-}                          & 7.18                & {-}                          \\ \hline
		\textbf{deeplab}                                       & \textbf{4.71}       & {\color[HTML]{009901} \textbf{43.80}}                & \textbf{4.52}       & {\color[HTML]{009901} \textbf{37.01}}                \\ \hline
		\textbf{unet}                                          & 6.51                & {\color[HTML]{009901} 22.29}                         & 6.80                & {\color[HTML]{009901} 5.22}                          \\ \hline
		\textbf{densenet}                                      & 7.24                & {\color[HTML]{009901} 13.59}                         & 7.20                & {\color[HTML]{FE0000} -0.30}                         \\ \hline
		\textbf{segnet}                                        & 6.48                & {\color[HTML]{009901} 22.61}                         & 6.64                & {\color[HTML]{009901} 7.51}                          \\ \hline
	\end{tabular}
	\label{tab_mae}
\end{table}

\subsubsection{Ablation study with training images}
\label{sec_image_ablation}

\begin{figure*}[!htbp]
	\includegraphics[width=0.245\textwidth]{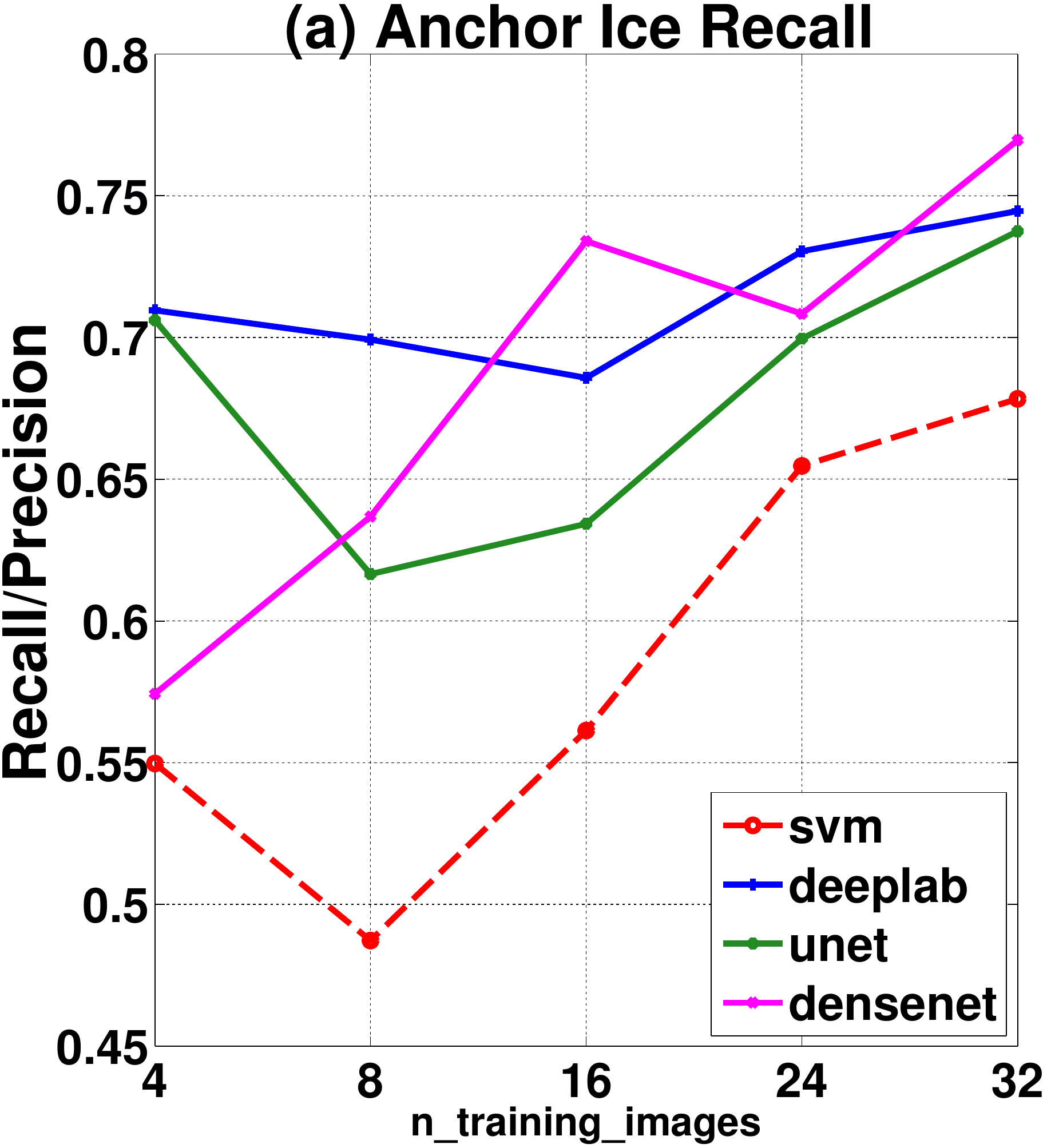}
	\includegraphics[width=0.245\textwidth]{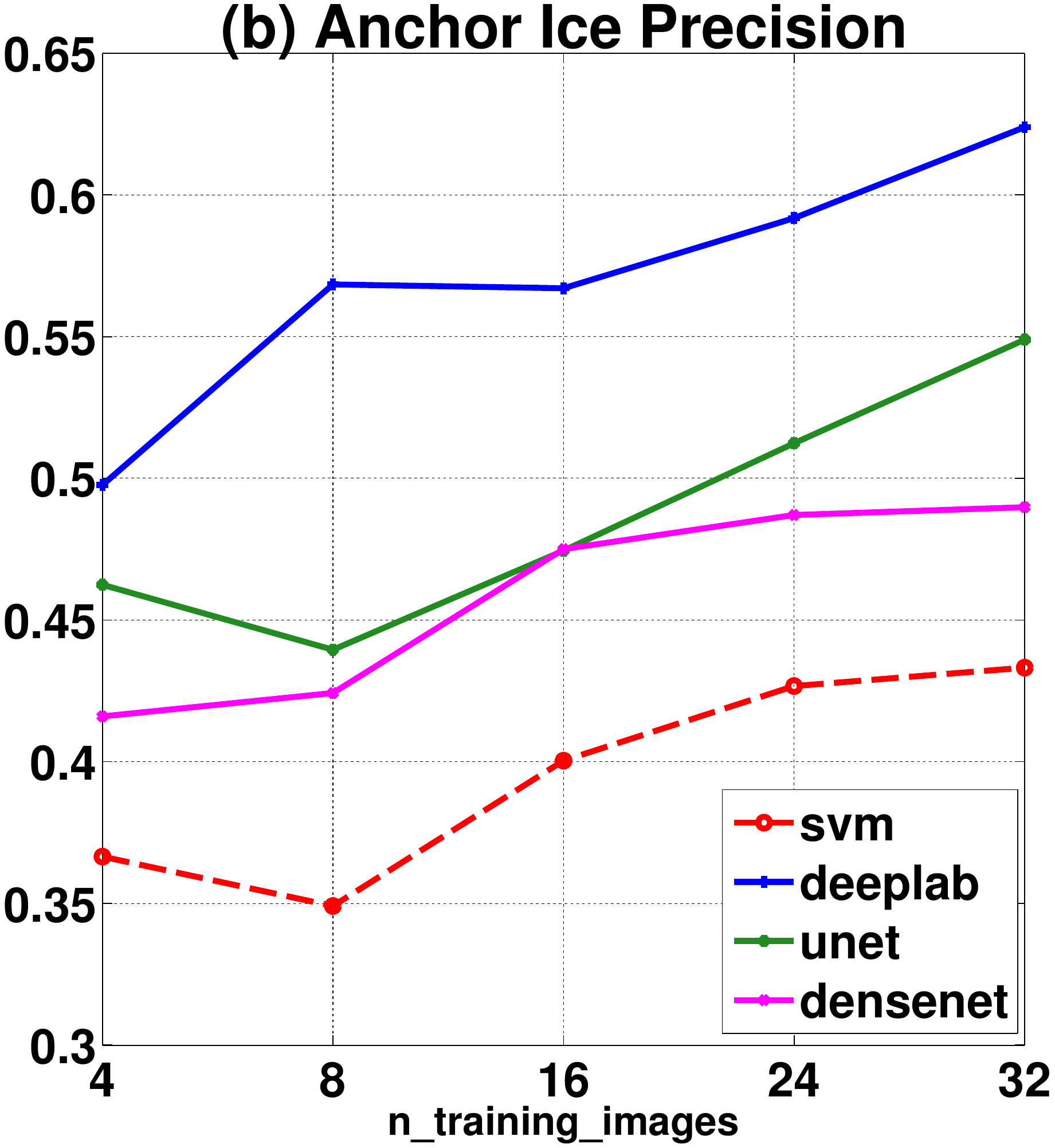}
	\includegraphics[width=0.245\textwidth]{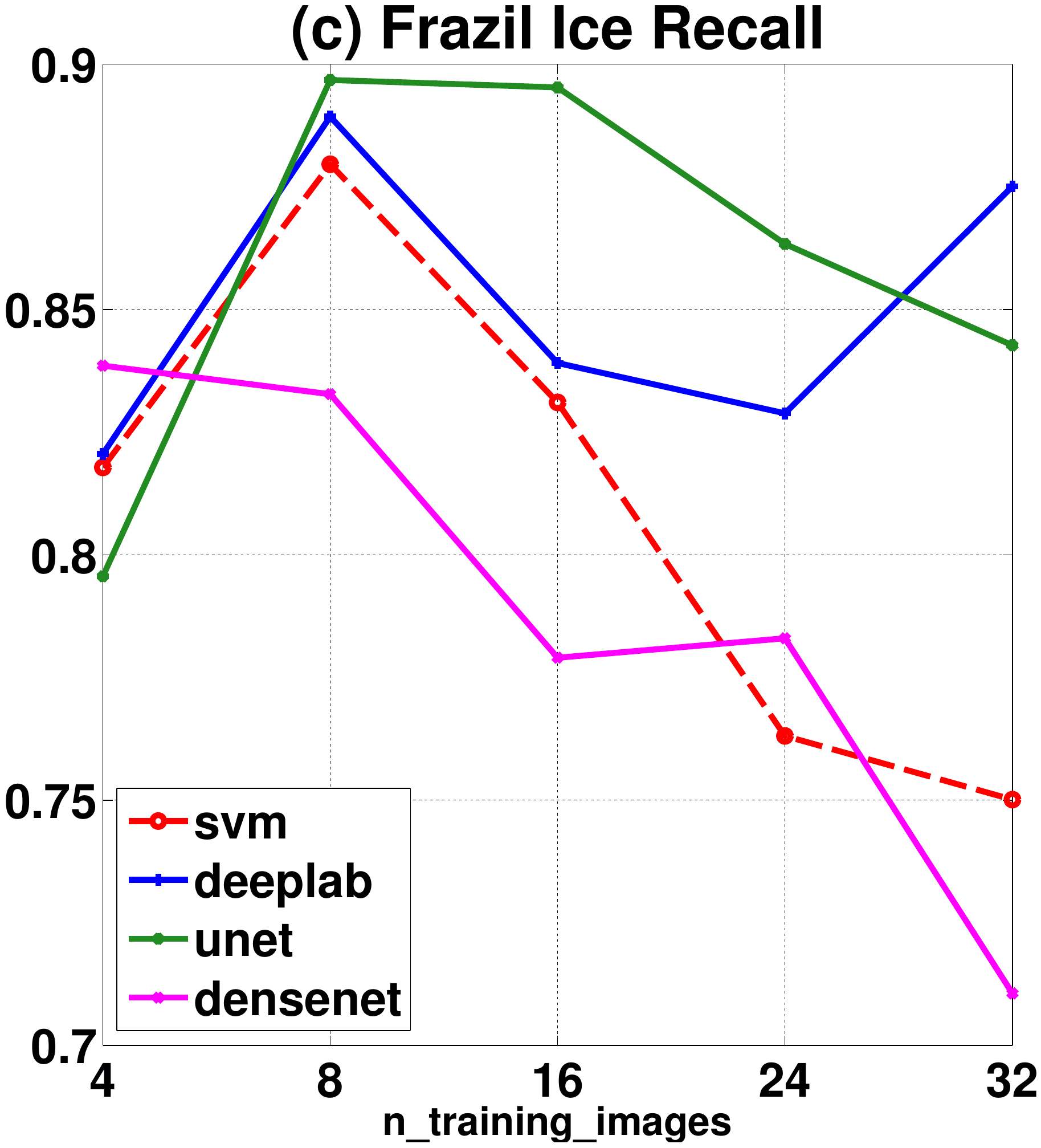}
	\includegraphics[width=0.245\textwidth]{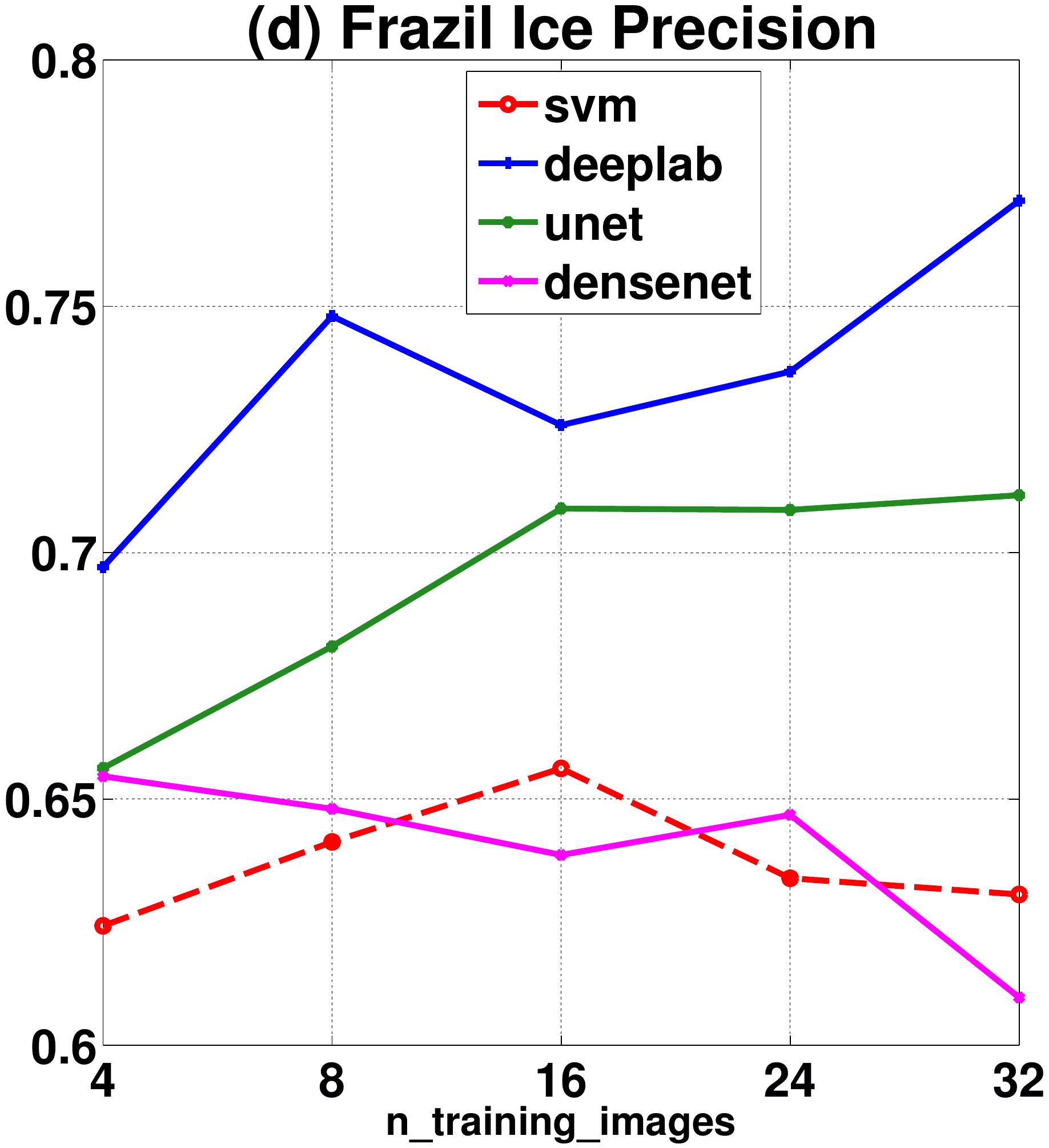}
	\caption{
		Results of ablation tests with training images for (a-b) anchor ice and (c-d) frazil ice.
		Note the variable Y-axis limits.
	}
	\label{fig_ablation_img}
\end{figure*}

\begin{figure*}[!htbp]
	\includegraphics[width=0.245\textwidth]{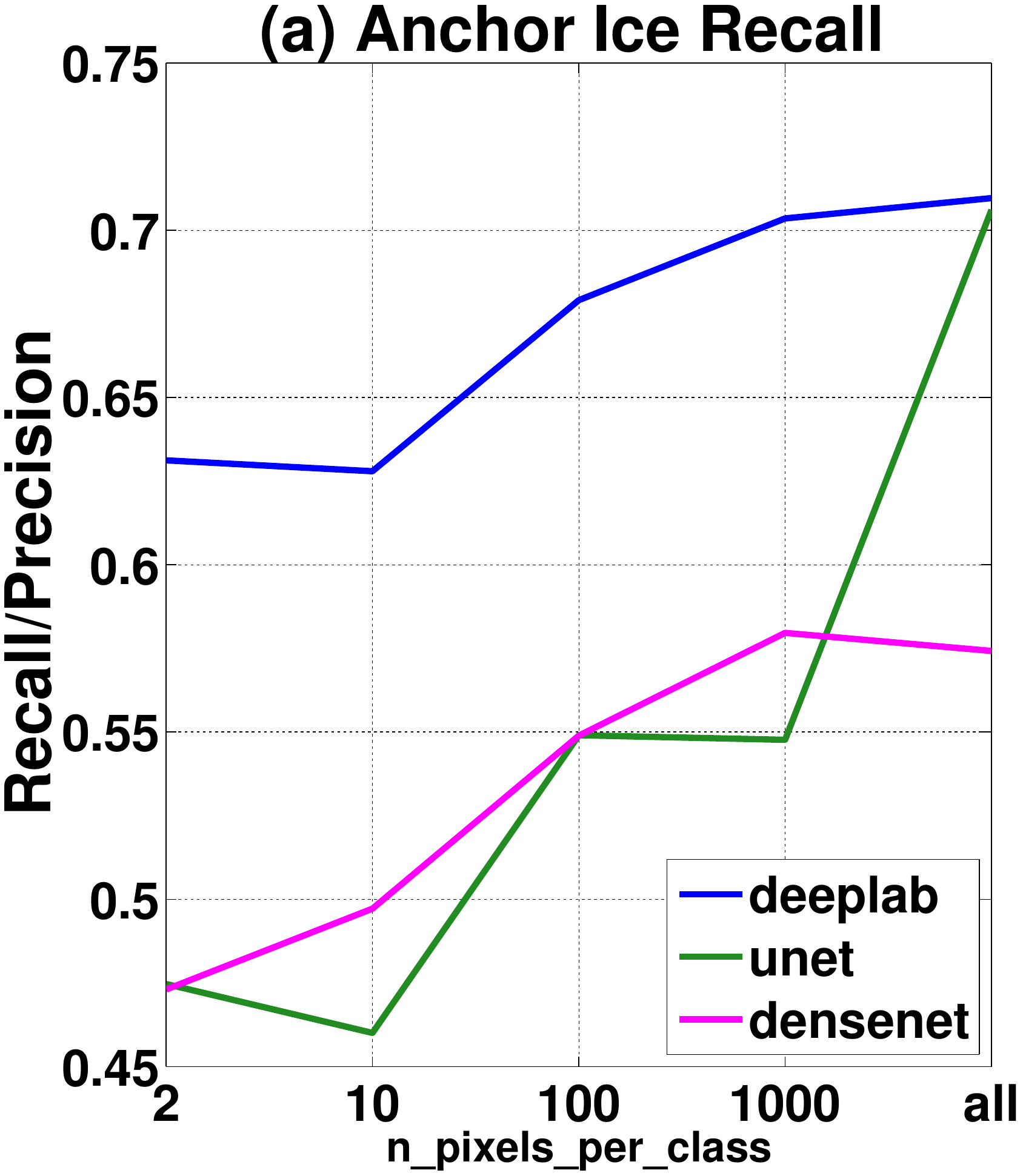}
	\includegraphics[width=0.245\textwidth]{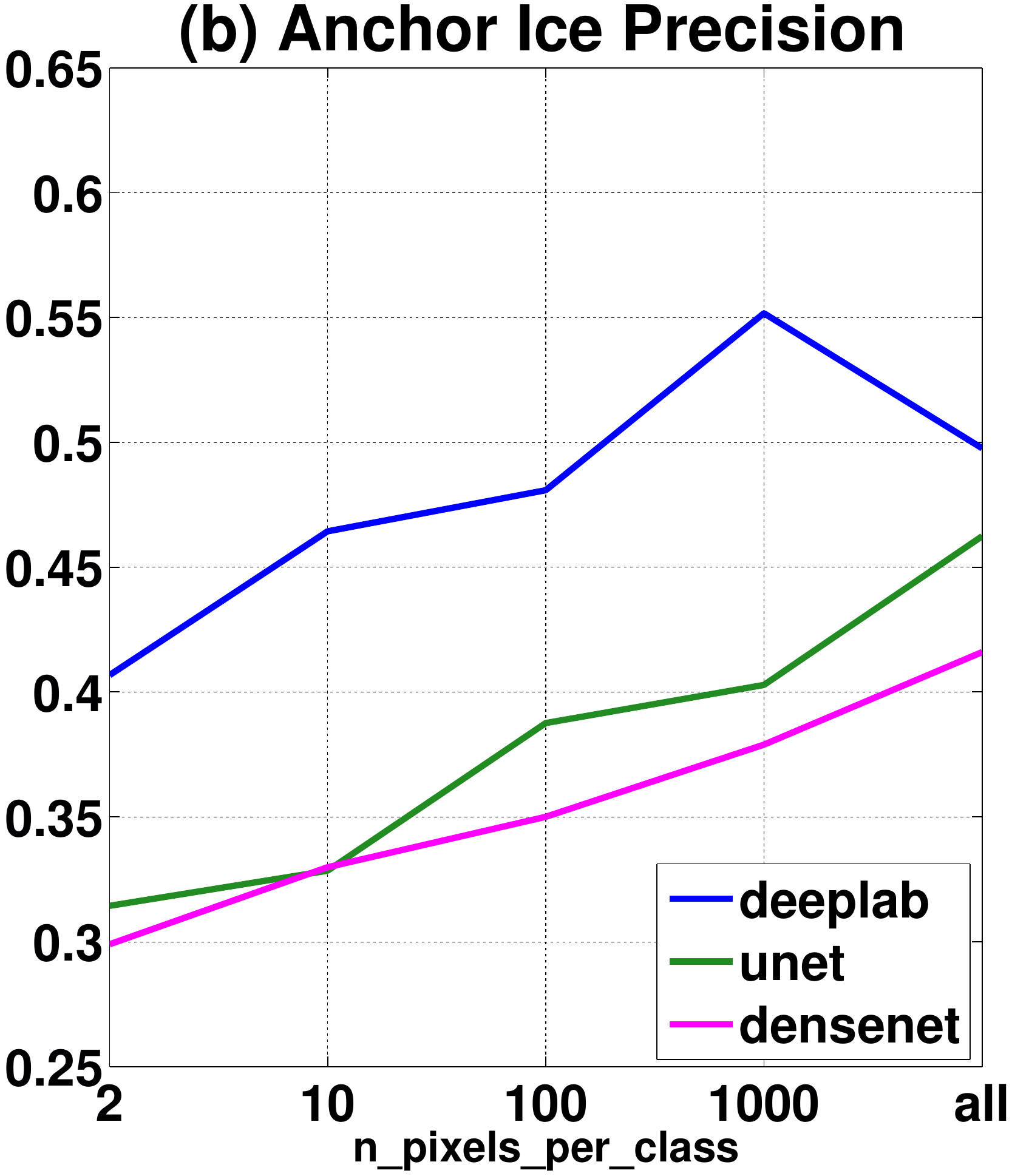}
	\includegraphics[width=0.245\textwidth]{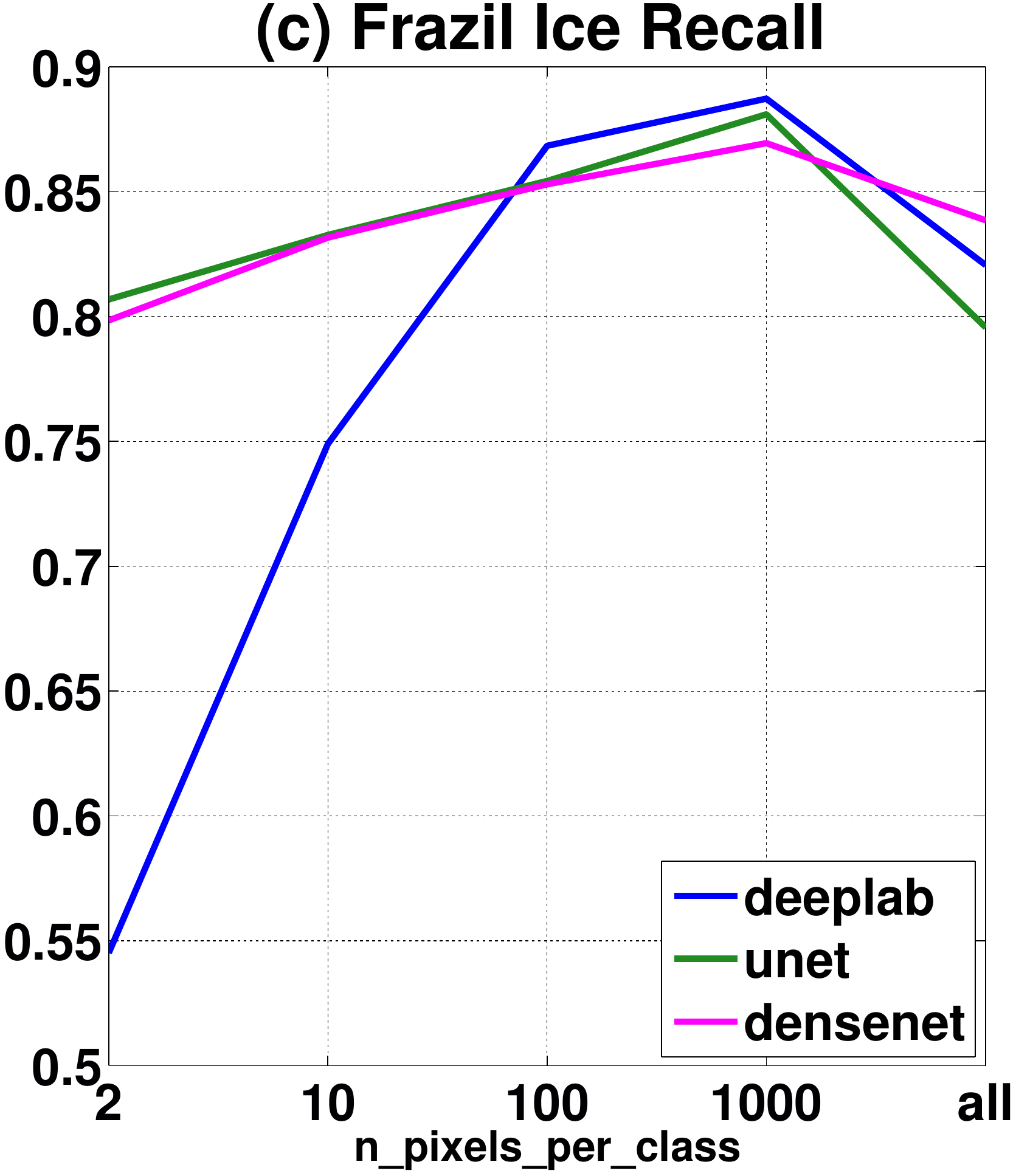}
	\includegraphics[width=0.245\textwidth]{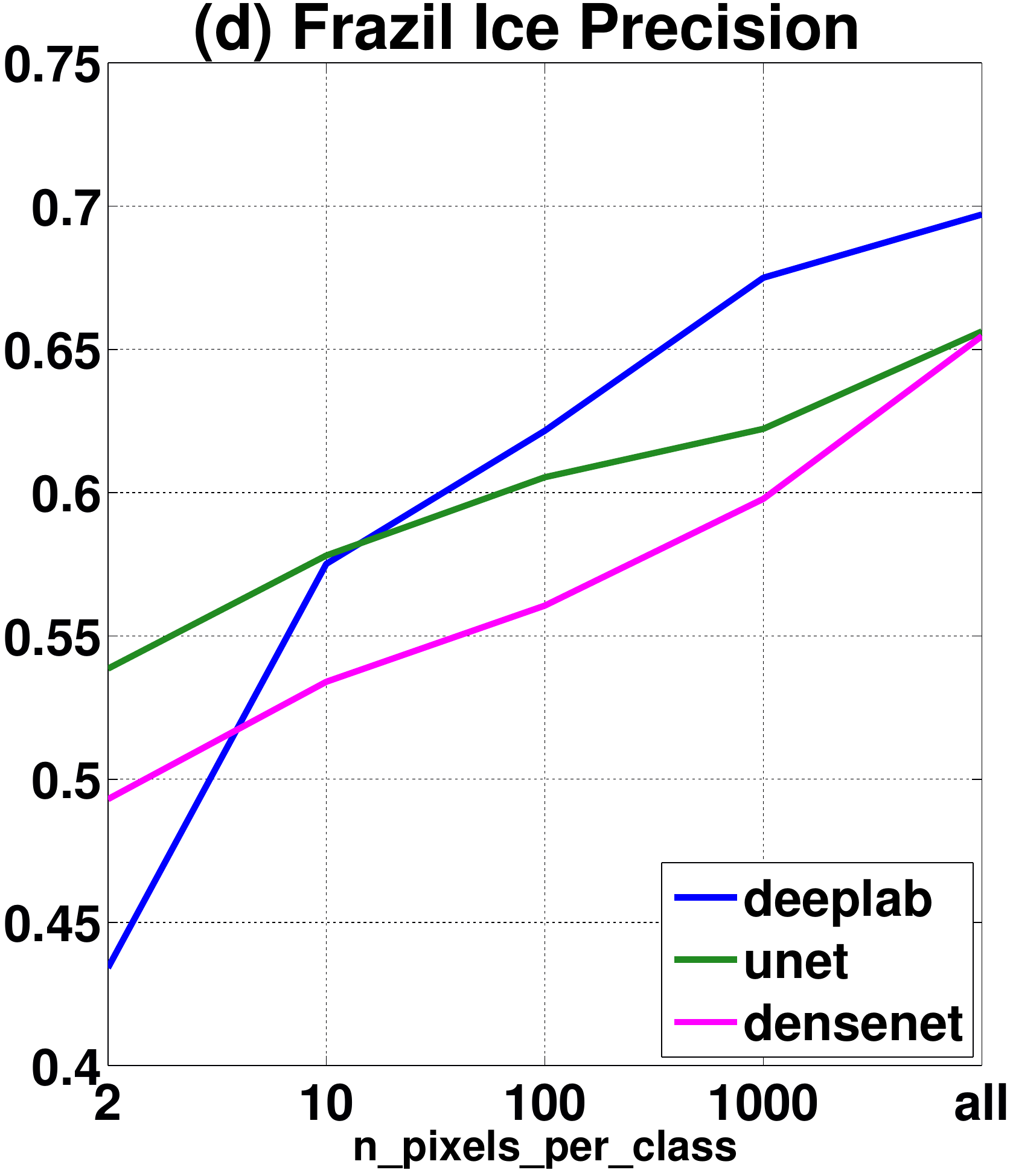}
	\caption{
		Results of ablation tests with selective pixels for (a-b) anchor ice and (c-d) frazil ice.
		Note the variable Y-axis limits.
	}
	\label{fig_ablation_pix}
\end{figure*}

For this study, models were trained using 4, 8, 16, 24 and 32 images and each one was tested using the same 18-image test set.
Results for both anchor and frazil ice are given in Fig. \ref{fig_ablation_img}.
Contrary to expectation, a distinct pattern of improvement with more images is not shown by most of the models.
There is a slight improvement in anchor ice performance but it seems too weak to clearly demonstrate model improvement with increase in training images.
%Also, experiments have shown that such trends are susceptible to the individual checkpoint that is used for testing since re-training some of these models for the same number of epochs or simply using a sufficiently different checkpoint from the same training session was found to change the performance significantly.
A more likely conclusion is that the test set is just too similar to the training set and does not contain enough challenging variation to allow the extra information from more training images to be reflected in the performance numbers.
This is lent some credence by the fact that the 50 labeled images were specifically chosen for their ease of labeling owing to the highly tedious and time-consuming nature of performing pixel-wise segmentations in addition to the high degree of ambiguity and subjectivity in classifying ice in the more difficult cases.
Combined with the fact that they were labeled by the same person, it would not be unusual for them to be similar, both in terms of content and level of challenge.

The inverse relation between anchor and frazil ice recall that was observed in the previous section is apparent here too.
For instance, the plot lines for anchor and frazil ice (Fig. \ref{fig_ablation_img} (a),(c)) are virtually reflections of each other for all models including SVM except perhaps for Deeplab.
Deeplab itself exhibits nearly constant recall though with a clearer upward trend in precision. 
It is also overall the best model, as in the previous section.
SVM shows the strongest improvement in anchor ice recall, along with the corresponding decline in frazil ice while DenseNet does so among the deep models.
Also, Fig. \ref{fig_ablation_img} (a) and (b) illustrate the superiority of deep models over SVM for anchor ice recognition more clearly than Table. \ref{tab_acc_iou_32}.
Deeplab and UNet largely maintain an appreciable superiority over SVM  for frazil ice too (Fig. \ref{fig_ablation_img} (c), (d)), though the overall improvement there is less distinct.
%with the exception of a significant dip with 16 images which was probably due to the aforementioned susceptibility of these results to the specific checkpoint used for testing.
Finally, DenseNet does seem to be the worst performing deep model, especially for frazil ice, but its competitiveness is still noteworthy considering that it has over two orders of magnitude fewer parameters than the other models (Table \ref{tab_trainable_params}).

%Results comparing the ablation performance of different models for anchor and frazil ice are given in Fig. \ref{fig_ablation_img}.
%Not much of interest is apparent here except that the dominance of Deeplab over the other deep models is much less prominent compared to Fig. \ref{fig_acc_iou_32} and it is DenseNet which seems to be the overall best performer for anchor ice while UNet performs best for frazil ice.
%The competitive performance of DenseNet is noteworthy considering that it has over 2 orders of magnitude fewer parameters than the other models, though this might be at least partially attributable to the limited challenges available in the test set.
%Unfortunately, this smaller size does not translate into higher speed or lesser memory requirement as both of these lie somewhere between UNet and Deeplab.
%It does result in a much smaller model which might be more suitable for a mobile device with limited storage or when a large number of these models specialized for different tasks are to be available simultaneously and then chosen dynamically at runtime.

\subsubsection{Ablation study with selective pixels}
\label{sec_pixel_ablation}

This study was performed by training models using 2, 10, 100 and 1000 pixels per class selected randomly from each $ K\times K $ training patch.
The training set was generated from only 4 training images and not subjected to augmentation.
%(last column of table \ref{tab_dataset_size}).
%Further, only 4 training images were used and the corresponding sub-image sets were also not subjected to augmentation.
Also,  $ K=640 $ was used for all models including DenseNet to ensure that the number of training pixels remained identical for all of them.
Further, in an attempt to counteract the limited challenges available in the 18-image test set, these models were tested on all of the remaining 46 labeled images.
Finally, SVM was not included here because its super-pixel based method \cite{Kalke17_cripe} does not lend itself well to training using randomly selected pixels.

Results are given in Fig. \ref{fig_ablation_pix}.
%Note that the \textit{all} pixel results are for the augmented sub image test set generated from 4 images while all others are for the unaugmented test set.
It turns out that selective pixel training has surprisingly little impact on quantitative performance except perhaps in the case of UNet with anchor ice and DeepLab with frazil ice.
Though there is a more strongly marked upward trend in performance compared to training images (Fig. \ref{fig_ablation_img}), it is not as significant as would be expected.
The case of 2 pixels per class is particularly remarkable.
When combined with the fact that the unaugmented training set contained only 46 patch images, this training was done using only 92 pixels per class or 276 pixels in all.
This might be another indicator of the limited challenges available in the test set.
This is further confirmed by the qualitative results on videos (Sec. \ref{sec_qualitative_videos}) that show a much more strongly marked difference than would be inferred by these plots.

\begin{figure*}[!htbp]
	\includegraphics[width=0.245\textwidth]{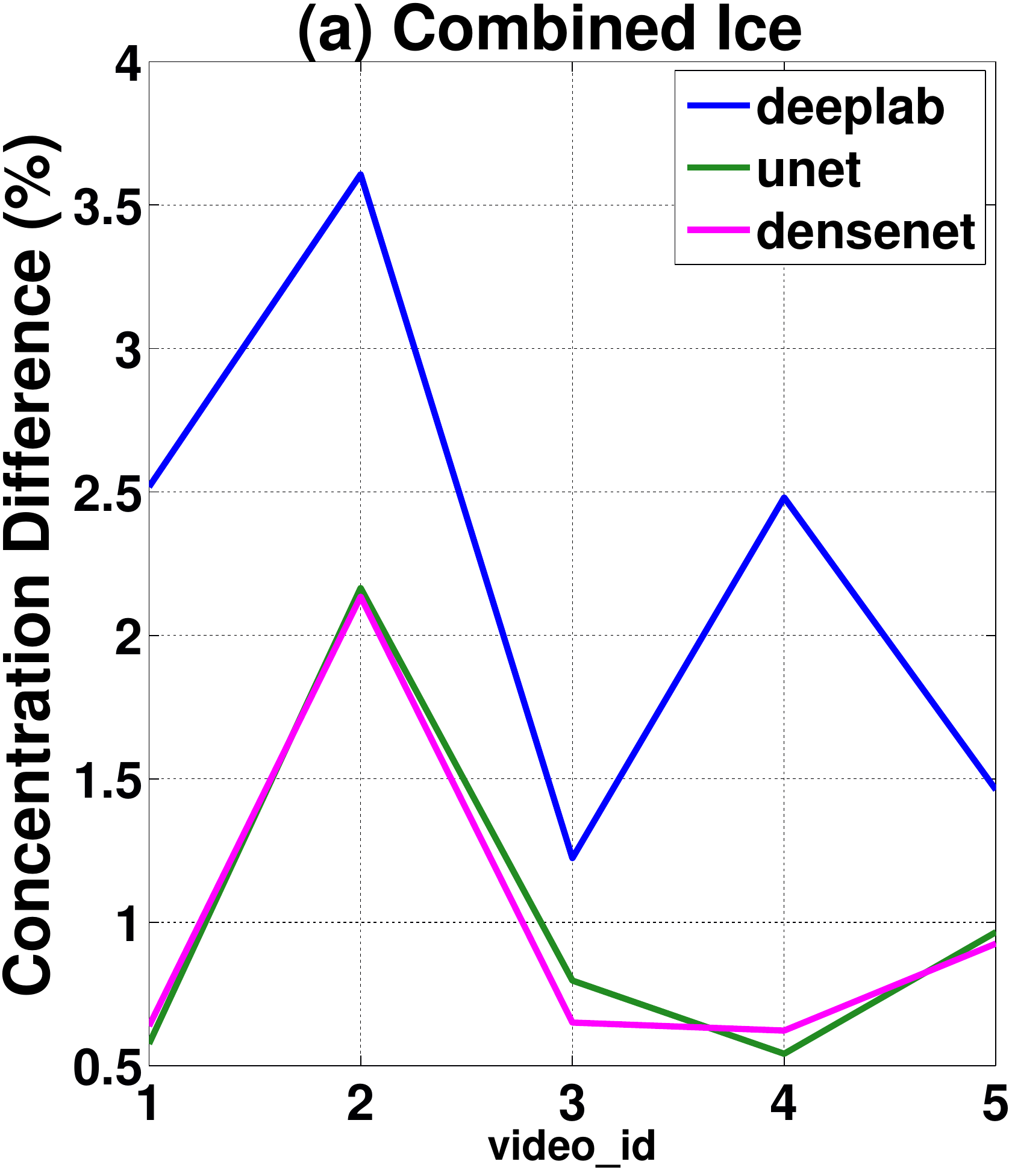}
	\includegraphics[width=0.245\textwidth]{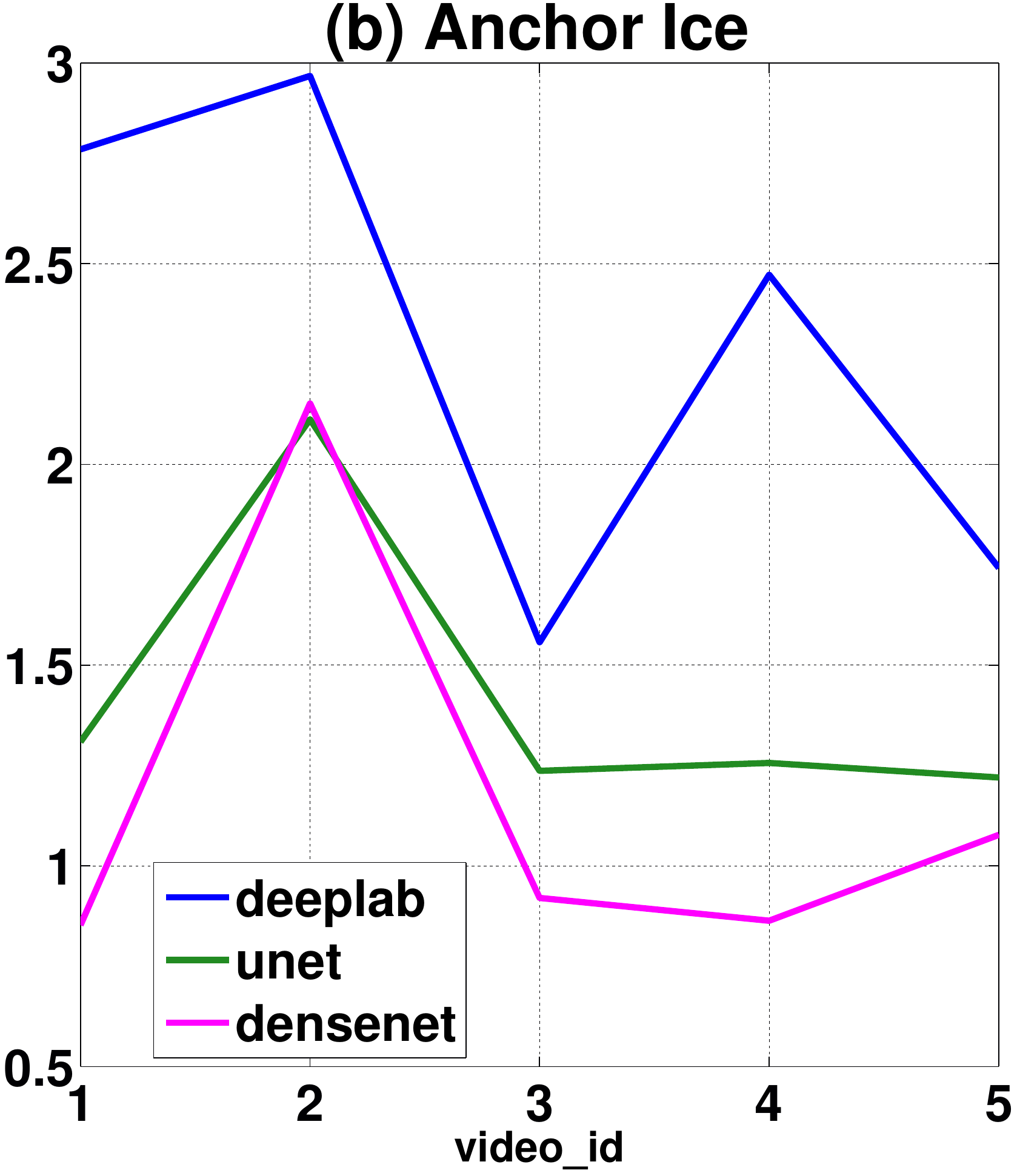}
	\includegraphics[width=0.245\textwidth]{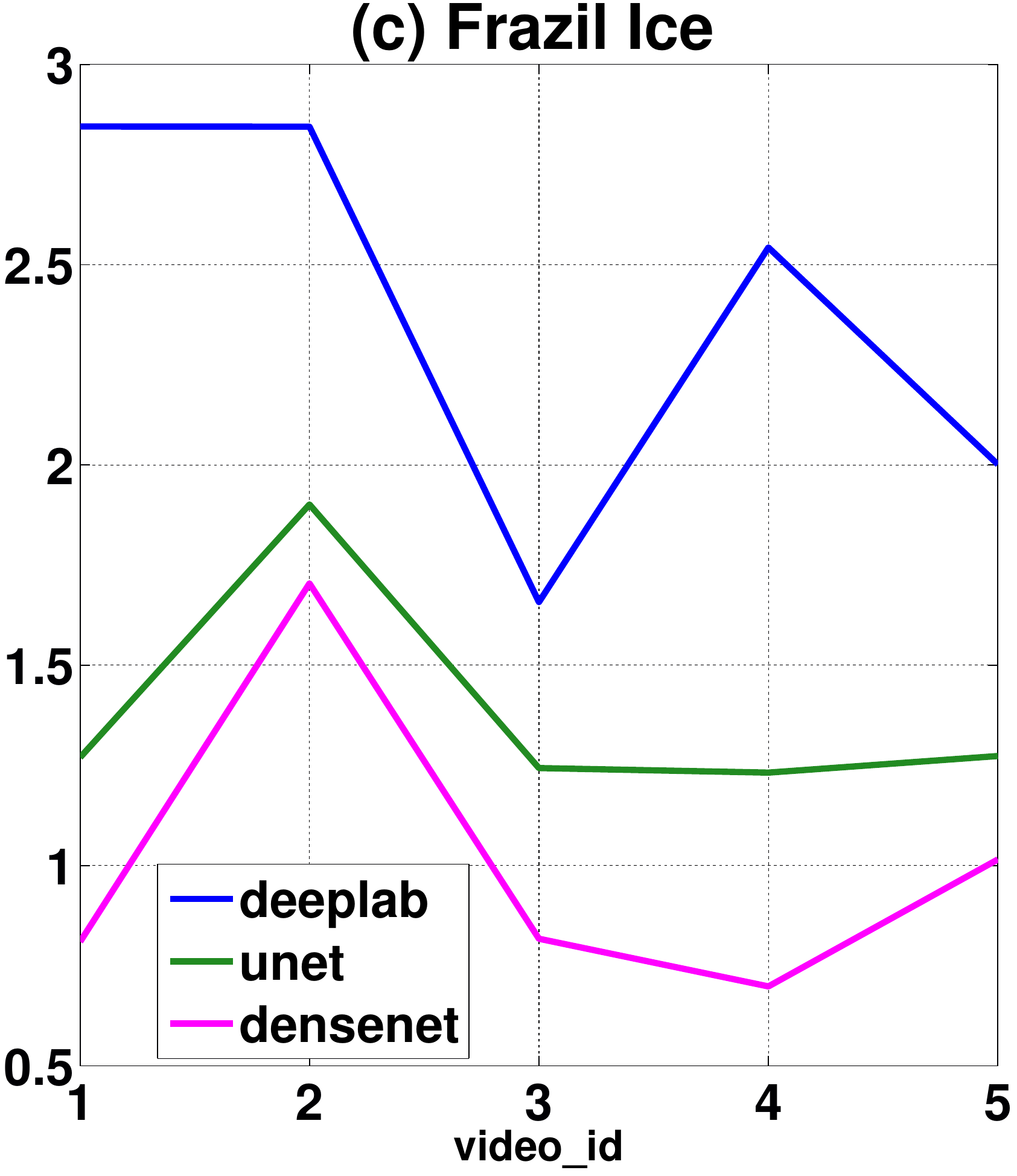}
	\includegraphics[width=0.245\textwidth]{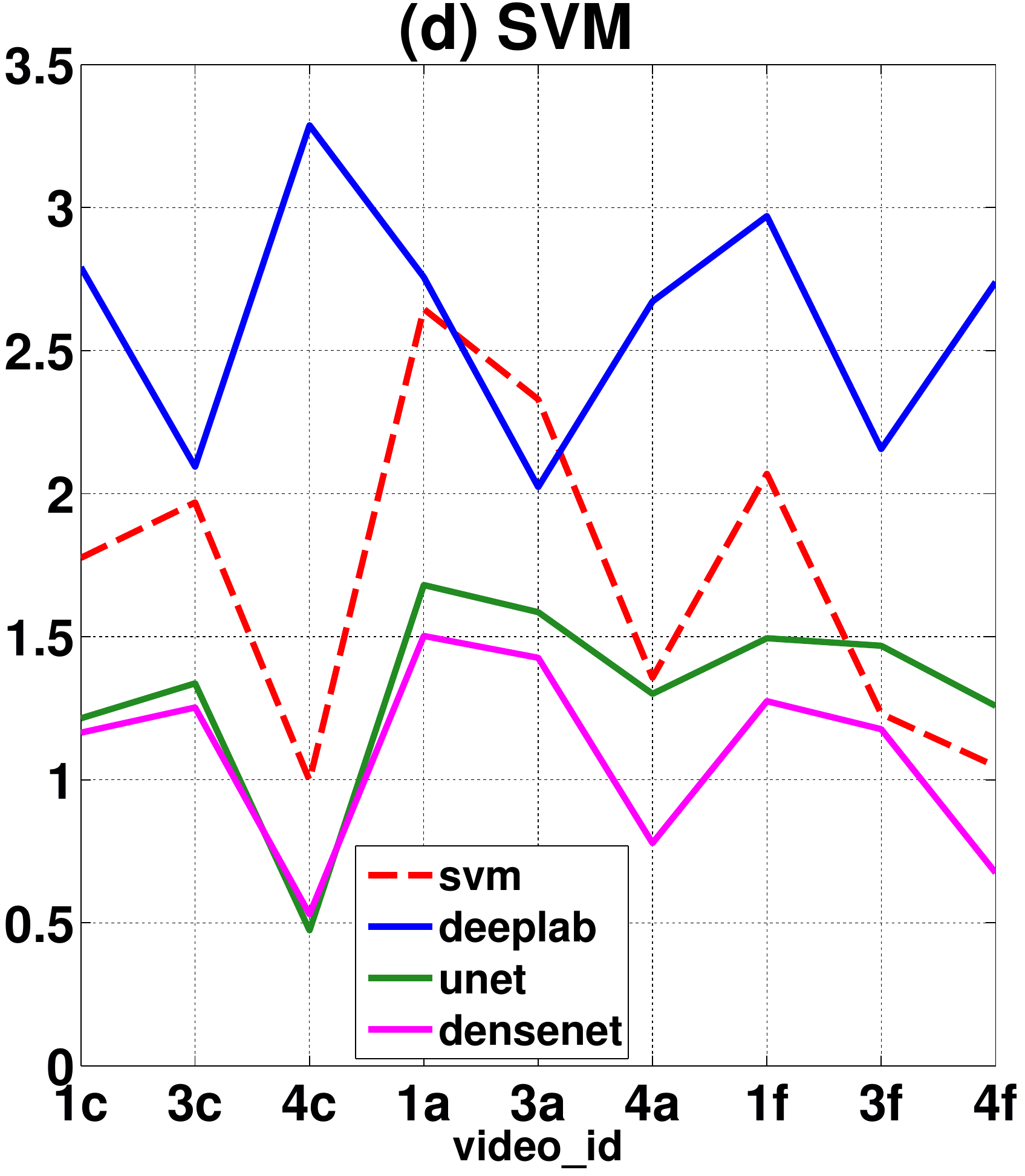}
	\caption{
		Mean ice concentration difference between consecutive frames for (a) both types of ice combined, (b) anchor ice, (c) frazil ice and (d) all three for SVM sequences.		
		Details of videos corresponding to the IDs are in the supplementary and suffixes \texttt{c}, \texttt{a} and \texttt{f} in (d) respectively refer to combined, anchor and frazil ice.
%		Note that the sequences tested for (d) consist of a subset of frames used for (a)-(c).
		%		so their results cannot be directly compared.
	}
	\label{fig_ice_conc_diff}
\end{figure*}

%\smallskip
\subsection{Qualitative Results}
\label{sec_qualitative}

\subsubsection{Images}
\label{sec_qualitative_images}

\begin{figure*}[!htbp]
	\includegraphics[width=\textwidth]{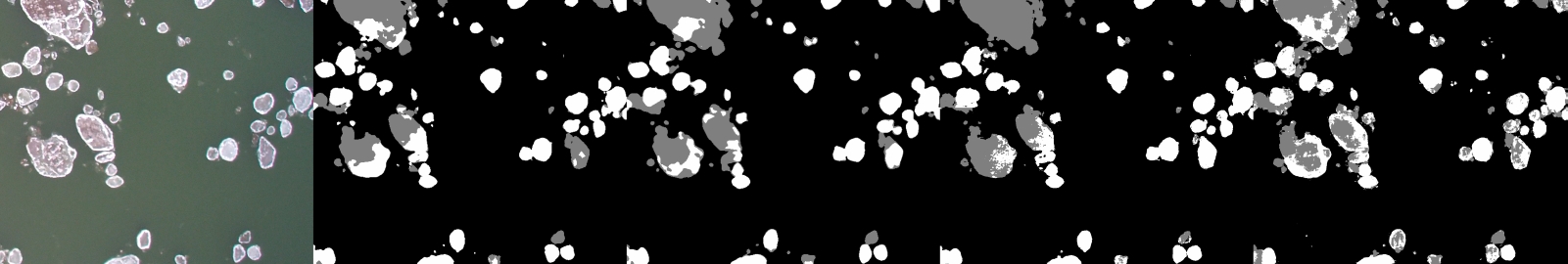}
	\includegraphics[width=\textwidth]{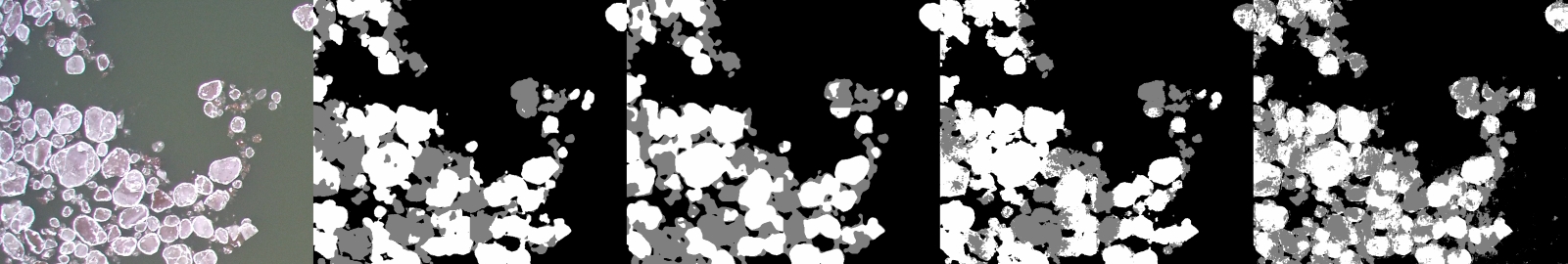}
	\includegraphics[width=\textwidth]{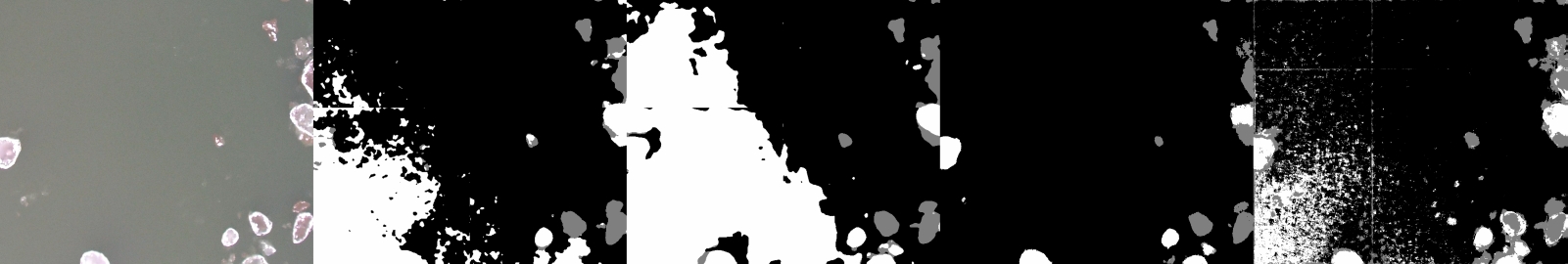}
	\includegraphics[width=\textwidth]{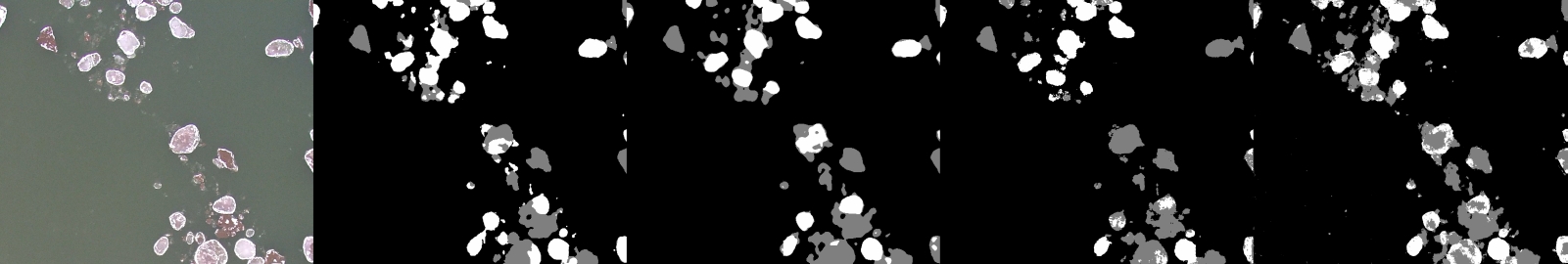}
	\includegraphics[width=\textwidth]{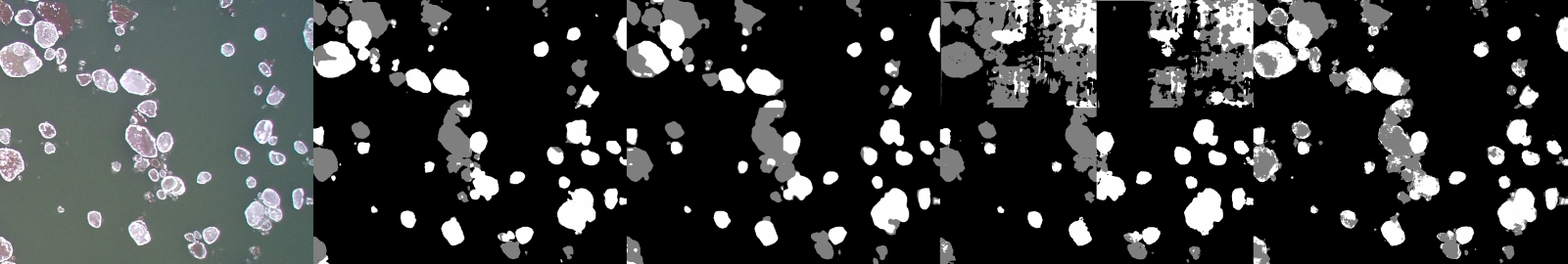}
	\includegraphics[width=\textwidth]{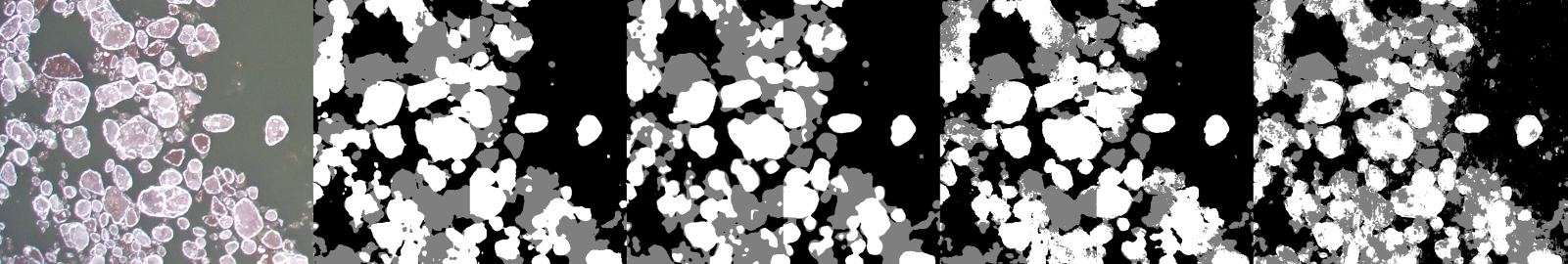}
	\includegraphics[width=\textwidth]{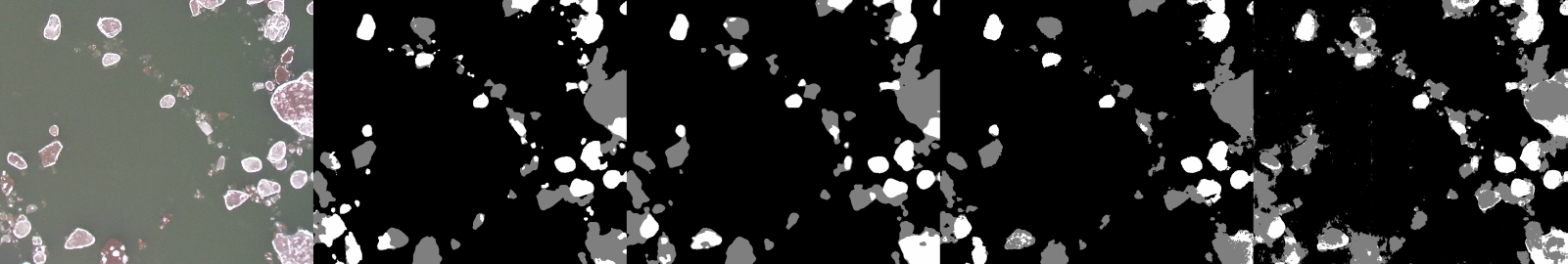}
	\caption{Results of testing the deep models on unlabeled images: left to right: raw image, UNet, SegNet, Deeplab, DenseNet}
	\label{fig_qualitative2}
\end{figure*}

Fig. 
%\ref{fig_qualitative1}, 
\ref{fig_qualitative2}
%and \ref{fig_qualitative3}
shows the results of applying the optimal configurations of the four models (as used in Sec. \ref{sec_summary}) to segment several images from the unlabeled test set.
Additional results are in the supplementary.
%\footnote{
%	(
%	\href{https://photos.app.goo.gl/TcK4tdsRmSmHxAKX9}{this} Google Photos album
%	).
%}
%UNet, SegNet and Deeplab were tested using $ K = 640  $ while Densenet used $ K=800 $.
Several interesting observations can be made.
First, both UNet and SegNet misclassify water as frazil ice in several cases where water covers most of the image, e.g. in 
%images 2 and 3 respectively and  of Fig. \ref{fig_qualitative1} and \ref{fig_qualitative2}.
image 3.
DenseNet too seems to be susceptible to this issue, albeit to a much lesser extent, though a careful examination of its video results (Table \ref{fig_ice_conc_diff}) shows this problem to be more prevalent than the images alone indicate.

Second, Deeplab results show the largest degree of discontinuity between adjacent patches due to its tendency to occasionally produce completely meaningless segmentations on some individual patches.
%Examples include image 6 in Fig. \ref{fig_qualitative1}, image 5 in Fig. \ref{fig_qualitative2} and images 1 and 4 in Fig. \ref{fig_qualitative3}.
Image 5 is an example.

Third, consistent with the quantitative results of the previous section, DenseNet is overall the best performing model, even though its results are slightly more fragmented than the others.
This is particularly noticeable in the more difficult cases of distinguishing between frazil and anchor ice when they both form part of the same ice pan.
Images 1 and 7 are examples.
%in Fig. \ref{fig_qualitative2} and image 1 in  Fig. \ref{fig_qualitative3}.

Qualitative results on labeled test image are available in the accompanying data \cite{ebax-1h44-19} as well as in Google Photos albums whose categorized links are given in the supplementary for convenience.

%Qualitative results on labeled test images are available in Google Photos albums whose links are in Table \ref{tab_qualitative_img_pix}.
%Each row in these albums shows (from left to right) the original image, its ground truth label and its segmentation result.

\subsubsection{Videos}
\label{sec_qualitative_videos}

%\begin{figure*}[!htbp]
%	\includegraphics[width=\textwidth]{images/stitched/img_18.jpg}
%	\includegraphics[width=\textwidth]{images/stitched/img_19.jpg}
%	\includegraphics[width=\textwidth]{images/stitched/img_20.jpg}
%	\includegraphics[width=\textwidth]{images/stitched/img_21.jpg}
%	\includegraphics[width=\textwidth]{images/stitched/img_8.jpg}
%	\includegraphics[width=\textwidth]{images/stitched/img_9.jpg}
%	\includegraphics[width=\textwidth]{images/stitched/img_10.jpg}
%	\caption{Results of applying the best configurations of the four models on unlabeled images: left to right: raw image, UNet, SegNet, Deeplab, DenseNet}
%	\label{fig_qualitative3}
%\end{figure*}

All the deep models were evaluated on 1 to 2 minutes sequences from 5 videos captured on 3 different days and containing wide variations in the scale and form of ice pans.
SVM took 5 minutes to process each frame so could only be evaluated on 30 seconds of video 1 and 10 seconds each of videos 3 and 4.
Also, selective pixel models were only evaluated on videos 1 and 3.
Results for all videos are available in the accompanying data \cite{ebax-1h44-19}.
The supplementary provides details of the tested sequences along with categorized links for some of the results.

The most noticeable point in these results is that Deeplab is susceptible to completely misclassifying individual randomly distributed patches which can lead to strong discontinuities when these patches are stitched together to create complete frames and the corresponding video.
This is quantitatively confirmed by Fig. \ref{fig_ice_conc_diff} that shows the mean ice concentration difference between consecutive frames in all of the videos for each of the two types of ice as well as both combined.
It can be seen there that Deeplab has significantly higher values than both UNet and DenseNet, being more than 3 times higher in several cases, while DenseNet almost always has the smallest difference, thus indicating the most consistent segmentation results.
Apart from confirming the limited challenges available in the labeled test sets, this inversion of relative performance between the 3 models as compared to the quantitative results in Sec. \ref{fig_ablation_img}, shows that DenseNet is able to generalize to new scenarios much better than Deeplab which has a tendency to overfit to the training images while UNet provides a good balance between generalization and overfitting.
Fig. \ref{fig_ice_conc_diff} (d) shows the concentration differences over the 3 sequences on which SVM was also tested.
SVM classifies pixels in groups of super pixels so it doesn't suffer from the issue of misclassifying entire patches that the deep models are susceptible to but, even so, its generalization ability is poor enough to give significantly higher concentration differences as compared to both DenseNet and UNet, though being slightly better than Deeplab.
Moreover, this group classification technique has the disadvantage of giving up blocky appearance to its segmentation masks, whose boundaries are often too coarse to correspond well with the actual ice pans.

Examination of the ablation test videos gives another important result that was mentioned in Sec. \ref{sec_pixel_ablation} - selective pixel training has significantly greater impact in practice than indicated by Fig. \ref{fig_ablation_pix}.
The segmentation masks seem to become more grainy and sparse as the number of pixels is decreased and there is a very noticeable difference between using 2 and 1000 pixels.
Similarly, a greater difference is apparent between the results produced by models trained on 4 and 32 images than suggested by Fig. \ref{fig_ablation_img}.

%Qualitative results on labeled test images are available in the following Google Photos albums:

%	\item UNet:
%		\href{}{32 images},
%		\href{}{24 images},
%		\href{}{16 images},
%		\href{}{8 images},
%		\href{}{4 images}

%
%
%\begin{itemize}
%	\item SVM:
%		,
%		,
%		,
%		,
%		
%		
%	\item Deeplab:
%		\href{https://photos.app.goo.gl/wr2Zjtb9ibYbfi7F8}{32}
%		\href{https://photos.app.goo.gl/tFjMLB83cuTdtSkM9}{24}
%		\href{https://photos.app.goo.gl/mu6SjdFn6WLL1R9Z9}{16}
%		\href{https://photos.app.goo.gl/HiDA4cb5wLL4bSWT8}{8}
%		\href{https://photos.app.goo.gl/vb1xHeH4mfYrHVPh7a}{4}
%
%	\item UNet:
%		\href{https://photos.app.goo.gl/VsWrwMKhzjbjD13e6}{32}
%		\href{https://photos.app.goo.gl/6Y9faGdUEERVfA3p6}{24}
%		\href{https://photos.app.goo.gl/DUWhRiTbZdq7gLev9}{16}
%		\href{https://photos.app.goo.gl/mq7MR1rqHwmfptPu6}{8}
%		\href{https://photos.app.goo.gl/7DxTHBBtBorg4SRC9}{4}
%		
%	\item SegNet:
%		\href{https://photos.app.goo.gl/5BRm8N1e817VGYGS7}{32}
%		\href{https://photos.app.goo.gl/jCZog4tU1TTa7Lyt6}{24}
%		\href{https://photos.app.goo.gl/wPZJrcWCYiVRWQKt5}{16}
%		\href{https://photos.app.goo.gl/1TxQtFYGd2vVTKzR6}{8}
%		\href{https://photos.app.goo.gl/RznSYPC1JbrAmkwc7}{4}
%		
%	\item DenseNet:
%		\href{https://photos.app.goo.gl/rPA4mWiMUEAz5sLz6}{32}
%		\href{https://photos.app.goo.gl/YEsgLLc9ywdbVpYZ9}{24}
%		\href{https://photos.app.goo.gl/ux1szzfLMaRyp2Sv8}{16}
%		\href{https://photos.app.goo.gl/xKC7xzBw9ZYwsrw88}{8}
%		\href{https://photos.app.goo.gl/9NdbTipG4mumUzhy8}{4}
%		
%	\item Deeplab/4 images:
%		\href{https://photos.app.goo.gl/sbWTZEfKQB26YqPM6}{1000 pixels},
%		\href{https://photos.app.goo.gl/nZq6uhUqCtmoPFrP9}{100 pixels},
%		\href{https://photos.app.goo.gl/6SiNbp6LqGGF3wo19}{10 pixels},
%		\href{https://photos.app.goo.gl/Qtmx1Rpp5ThK1yjj6}{2 pixels}
%		
%	\item UNet/4 images:
%		\href{https://photos.app.goo.gl/CtRUqXhF136ZisRx9}{1000 pixels},
%		\href{https://photos.app.goo.gl/bmm6NNdSEUovuPj3A}{100 pixels},
%		\href{https://photos.app.goo.gl/jU7WpvWS3fjJ7sab7in}{10 pixels},
%		\href{https://photos.app.goo.gl/FkB2FwsnxariPowVA}{2 pixels}
%		
%	\item SegNet/4 images:
%		\href{https://photos.app.goo.gl/5tFX1fMuViPY5MzK7}{1000 pixels},
%		\href{https://photos.app.goo.gl/4LDwUsx8jP8kBJtH7}{100 pixels},
%		\href{https://photos.app.goo.gl/ipbtqnMriu6nuVaM7}{10 pixels},
%		\href{https://photos.app.goo.gl/hVmCLj4SLEmgrFHn8}{2 pixels}
%		
%	\item DenseNet/4 images:
%		\href{https://photos.app.goo.gl/KLBPaYjEAtFn75PS8}{1000 pixels},
%		\href{https://photos.app.goo.gl/QGBa3bKu19ZWADPs7}{100 pixels},
%		\href{https://photos.app.goo.gl/jU7WpvWRUQS2bTXyaBgpXak8}{10 pixels},
%		\href{https://photos.app.goo.gl/RJ6He9MiWkYw546y9}{2 pixels}
%		
%
%%	\item UNet/100 pixels: \href{https://photos.app.goo.gl/bmm6NNdSEUovuPj3A}
%%	\item UNet/10 pixels:  \href{https://photos.app.goo.gl/jU7WpvWS3fjJ7sab7in}
%%	\item UNet/2 pixels:  \href{https://photos.app.goo.gl/FkB2FwsnxariPowVA}
%%	\item DenseNet/1000 pixels: \href{https://photos.app.goo.gl/KLBPaYjEAtFn75PS8}
%%	\item DenseNet/100 pixels: \href{https://photos.app.goo.gl/QGBa3bKu19ZWADPs7}
%%	\item DenseNet/10 pixels: \href{https://photos.app.goo.gl/RUQS2bTXyaBgpXak8}
%%	\item DenseNet/2 pixels: \href{https://photos.app.goo.gl/RJ6He9MiWkYw546y9}	
%\end{itemize}

\section{Conclusions and Future Work}
\label{sec_conclusions}
This paper presented the results of using four state of the art deep CNNs for segmenting images and videos of river surface into water and two types of ice.
Three of these - UNet, SegNet and Deeplab - are previously published and well studied methods while the fourth one - DenseNet - is a new method, though based on an existing architecture.
All of the models provided considerable improvements over previous attempts using SVM.
These were particularly significant for the most challenging case of anchor ice where around $ 20\% $ increase in both recall and precision were obtained by the four models combined while the single best model - Deeplab - provided respective improvements of $ 13 - 19\%  $ in absolute and $ 21 - 44\% $ in relative terms.
Frazil ice performance was slightly less impressive but still surpassed SVM by $ 12 - 14\% $ in absolute and  $ 16 - 22\% $ in relative terms.
Significant improvements were obtained in ice concentration estimation accuracy too, with Deeplab providing around $ 3\% $ absolute and $ 40\% $ relative decrease in MAE over SVM for both types of ice.
%increase in accuracy over, especially in distinguishing between the two types of ice - nearly $ 20\% $ for anchor ice and $ 15\% $ for frazil ice.

Among the four models, Deeplab gave the best quantitative performance on the labeled test set - $ 5-10\% $ improvement in precision with similar recall for anchor ice and $ 3-15\% $ better recall and precision for frazil ice - but showed poor generalization ability by giving the worst qualitative results on the unlabeled images and videos - up to 7-fold mean concentration difference on the videos.
DenseNet, on the other hand, gave poor quantitative results but demonstrated excellent generalization ability on the unlabeled data.
UNet provided a good balance between the two and might be taken to be the single best model tested here, if such a one needs to be chosen.
%This provides a promising avenue for future exploration that might be able to yield much better performance with more layers and training images.
Finally, this paper also demonstrated reasonable success in handling the lack of labeled images using data augmentation.
Further improvements in this direction would be obtained by using the augmented dataset as the starting point for a semi-automated boot-strapping process where segmentation results on unlabeled images would be corrected manually to yield more labeled data to be used for training better models.
%This involves training successively better models by manually correcting the segmentation results produced on the test images by each stage of the process and adding these corrected images to the labeled set for training the next stage model.
% Another solution for this problem might be offered by recent methods for real time video segmentation that are able to deal with limited labeled data by using motion information \cite{}.

%\clearpage
{\small
\bibliographystyle{ieee}
\bibliography{references}
}

\end{document}